%% file: manuscript.tex
\documentclass[]{jair}
\usepackage{longtable}
\usepackage{array}
\usepackage{float}
\floatstyle{plaintop}
\newfloat{algorithm}{htbp}{loa}
\floatname{algorithm}{Algorithm}

\setcopyright{cc}
\copyrightyear{2026}
\JAIRAE{Reshef Meir}
\JAIRTrack{}
\input{publication_metadata.tex}

\RequirePackage[
  datamodel=acmdatamodel,
  style=acmauthoryear,
  backend=biber,
  giveninits=true,
  uniquename=init
]{biblatex}

\begin{document}
\title[Semi-Strongly Solved]{Semi-Strongly Solved: a New Definition Leading Computer to Perfect Gameplay}
\author{Hiroki Takizawa}
\orcid{0000-0002-2837-0690}
\email{contact@hiroki-takizawa.name}
\affiliation{%
  \institution{Preferred Networks Inc.}
  \city{Chiyoda-ku}
  \state{Tokyo}
  \country{Japan}
}
\renewcommand{\shortauthors}{Takizawa}

\begin{abstract}
Strong solving certifies optimal play from every reachable position, but its state-space coverage can be prohibitive. Weak solving establishes the initial value and a strategy that guarantees it against arbitrary replies; it need not determine exact values or locally optimal responses at every position reached after an opponent's mistakes. We introduce \emph{semi-strong solving}, which certifies exact values and canonical optimal moves on a region \(R\): positions reachable when one fixed player follows a canonical optimal policy throughout play and the opponent may move arbitrarily, taking the union over both player orientations. A fixed tie-breaking rule specifies the canonical policy. Our \emph{reopening alpha--beta} framework extends alpha--beta/PVS with explicit value and move-coverage obligations. Under ideal move ordering on a uniform game tree, it visits \(O(d\,b^{d/2})\) nodes for fixed branching factor \(b\) and move depth \(d\), including auxiliary proof work. For \(6\times6\) Othello, we construct a semi-strong solution artifact, whereas an attempted exhaustive enumeration exceeds \(4\times10^{12}\) positions before exhausting storage. For \(7\times6\) Connect Four, an experiment using an existing strong solution as an exact-value oracle yields a semi-strong certification structure with \(9{,}074\) times fewer positions than the published rule-reachable baseline under matched counting conventions. This ratio measures structural coverage under oracle ordering, not end-to-end computation. Semi-strong solving provides exact local play within a specified region while making the additional certification cost explicit.
\end{abstract}

\maketitle

\section{Introduction}
\label{sec:introduction}

Game-solving results differ in what they guarantee \cite{allis1994searching-thesis}.
An \emph{ultra-weak} solution establishes the initial game-theoretic value, possibly without providing a playable strategy; the strategy-stealing argument for Hex is a familiar example \cite{gale1979hex}.
A \emph{weak} solution also provides a strategy that guarantees the initial value against arbitrary opponent replies.
A \emph{strong} solution determines the value and optimal play at every position reachable by legal play.
Connect Four, Nine Men's Morris, and Checkers illustrate the achievements of game-solving research under these notions \cite{allis1994searching-thesis,boeck2025connect4strong,gasser1996solving-nine-mens-morris,schaeffer2007checkers}.

The distinction concerns more than the initial position.
A weak solution can contain a substantial proof tree, with exact values or sound bounds at its nodes, and its strategy already handles arbitrary opponent replies.
However, the guarantee of achieving the initial value does not generally require exploiting every later mistake as well as possible.
For example, a strategy that guarantees a draw from a drawn initial position can remain a valid weak solution even if it misses a win made available by an opponent's error.
A strong solution supplies the exact local value and an optimal response at that new position, but obtaining this information for every reachable position may be much more expensive.

We introduce \emph{semi-strong solving} to target the positions needed for locally optimal play against an unrestricted opponent.
Fix a tie-breaking rule among optimal moves, giving a canonical optimal policy.
For each choice of player as the \emph{optimal agent}, allow that player to follow this policy throughout the game and allow the \emph{free agent} to choose any legal move.
The union of the positions reached in these two orientations is the \emph{certified region} \(R\).
A semi-strong solution determines exact values on \(R\) and the canonical move at its optimal-agent decision points.
In human--AI play, the human is the free agent and the AI is the optimal agent. It suffices for the AI to follow its canonical policy to keep play in \(R\), however often the human makes mistakes.
Positions reached by other histories receive no additional guarantee unless they also belong to \(R\).
In particular, selecting a different equally optimal move need not preserve membership in the region fixed by the chosen tie-breaking rule.

\paragraph{An Informal Size Comparison.}
Table~\ref{tab:solving-overview} summarizes the guarantees and compares node counts under idealized assumptions.
The tree model has fixed branching factor \(b\ge2\), move depth \(d\), no transpositions, and ideal move ordering; parity effects are suppressed in the asymptotic notation.
The weak and semi-strong entries count nodes visited to construct a proof, including nodes that supply only bounds.
They are neither counts of positions with exact certified values nor universal complexity bounds for the solution notions.
An ultra-weak solution requires only one value, but there is no corresponding general bound of one search node.

\begin{table}[htbp]
\caption{Solving notions and illustrative search complexity under idealized assumptions. The semi-strong bound is proved in Section~\ref{sec:complexity}; the alpha--beta reference is \cite{knuth1975analysis-alphabeta}.}
\label{tab:solving-overview}
\centering
\begin{tabular}{>{\raggedright\arraybackslash}p{0.12\linewidth}>{\raggedright\arraybackslash}p{0.52\linewidth}>{\raggedright\arraybackslash}p{0.27\linewidth}}
\hline
Notion & Required guarantee & Idealized node count \\
\hline
Ultra-weak & Initial value & No general search bound \\
Weak & Initial value and a strategy guaranteeing it & \(\Theta(b^{d/2})\) with ideal alpha--beta \\
Semi-strong & Exact values on \(R\), plus canonical moves at optimal-agent turns & \(O(d\,b^{d/2})\) with reopening alpha--beta \\
Strong & Values and optimal play at all reachable positions & \(\Theta(b^d)\) for full tree enumeration \\
\hline
\end{tabular}
\end{table}

The intended use is an engine that plays from the initial position against a player who may make mistakes, with exact local value queries and responses throughout the supported play.
This can also support analysis of those games.
If only the perfect-play result is needed, weak solving is sufficient; if arbitrary starting positions must be analyzed, strong solving provides broader coverage.
Semi-strong solving addresses the intermediate requirement, and its usefulness depends on whether that additional coverage justifies its construction and storage costs.

We compute semi-strong solutions using \emph{reopening alpha--beta}, a node-kind-aware alpha--beta/PVS framework \cite{pearl1980scout,pearl1980asymptotic,reinefeld1983negascout}.
After the human departs from the canonical perfect-play line, reopening certifies the exact value and canonical optimal response at each subsequent AI turn, including after further human mistakes.
Under ideal move ordering, the canonical response is searched first, and its continuation retains the certification obligations for future human choices.
Each other AI move needs only an ordinary null-window alpha--beta search to show that it cannot improve on that response; the entire auxiliary search below it retains the classical cut/all structure \cite{fishburn1980nullwindow,knuth1975analysis-alphabeta}.
The \(O(d\,b^{d/2})\) idealized bound accounts for both the certified continuations and these auxiliary searches throughout the region; it is not a worst-case bound under arbitrary ordering.

\paragraph{Contributions.}
We define semi-strong solving and its deterministic certified region, prove the obligations and idealized node bound of reopening alpha--beta, and evaluate the resulting coverage on two games.
For score-valued \(6\times6\) Othello, we construct and release a solution artifact and auxiliary verification data.
For WDL-valued Connect Four, an exact-value oracle experiment isolates the effect of the certified-region requirement under ideal ordering.
The experiments report distinct-position counts and distinguish deployed information from auxiliary proof work.

\section{Preliminaries and Definitions}
\label{sec:preliminaries}

This section fixes the game model, the semi-strong setting, and the terminology that will be used throughout the paper.
In particular, we define the certified region $R$, the node kinds and their associated certification obligations, and the two output objects of our framework: a \emph{solution artifact} and a \emph{proof certificate}.
We also fix a deterministic tie-breaking convention, which is used to make the certified region and the produced artifacts canonical.

\subsection{Game Model and Game-Theoretic Value}
\label{sec:game-model}

We consider a two-player, zero-sum, perfect-information game with no chance nodes and alternating turns. A move, including a forced pass, changes the side to move.
A \emph{position} $p$ encodes the full game state including side-to-move and any rule state (e.g., repetition counters) if relevant.
Let $M(p)$ denote the set of legal moves at $p$ and let $p\cdot m$ denote the successor position obtained by applying $m\in M(p)$ to $p$.
We allow rule-mandated administrative moves such as passes to appear in $M(p)$.
In particular, in Othello, if the side to move has no ordinary placement move but the game is not terminal, then $M(p)$ consists of the unique forced-pass move.

A position $p$ is \emph{terminal} if the game ends at $p$ and no further moves are available.
We assume that each terminal position $p$ has a well-defined \emph{terminal utility} (payoff) given by $\texttt{value}(p)$.
The \emph{game-theoretic value} $V(p)$ of any (terminal or non-terminal) position is defined as the minimax value induced by this terminal utility:
\[
V(p)=
\begin{cases}
\texttt{value}(p), & \text{if $p$ is terminal},\\[2mm]
\max\limits_{m\in M(p)}\; -V(p\cdot m), & \text{otherwise},
\end{cases}
\]
where $V(p)$ is always interpreted from the perspective of the side to move at $p$ (negamax convention).
Different choices of terminal utility (e.g., WDL vs.\ score difference in Othello) therefore define different games in this sense and may induce different sets of optimal moves; our framework applies to any such specification.
We assume a finite game tree, so every play terminates and the recursive definition of $V(\cdot)$ and the inductions below are well-defined.

\paragraph{Remark on ties and canonical optimal moves.}
Unless stated otherwise, ties are allowed (multiple optimal moves may exist).
The value function $V(\cdot)$ is well-defined regardless of ties, but some parts of our framework (in particular, the definition of the certified region and the construction of a deterministic artifact) benefit from selecting a \emph{unique} optimal move at each position.

We therefore fix, once and for all, a deterministic tie-breaking rule $\mathrm{TB}$ that, at each non-terminal position $p$, selects one move from any nonempty subset of $M(p)$.
Examples include a fixed total order on move encodings or a deterministic, position-dependent heuristic order. For the canonical region to be a set of positions, this choice must be consistent whenever the same position is encountered; dependence on search history must be resolved by fixing the selected policy.
For every non-terminal position $p$, we define the \emph{canonical optimal move}
\[
m^{\star}(p)\;:=\;\mathrm{TB}\!\left(\arg\max_{m\in M(p)} -V(p\cdot m)\right).
\]
This convention never changes $V(\cdot)$; it only selects a unique representative among optimal moves.
Throughout the paper, whenever we refer to ``the'' optimal move (for purposes of defining reachability, PV selection, and storing a best move in an artifact), it means this canonical choice $m^{\star}(p)$.

\subsection{Semi-Strong Setting and the Certified Region}
\label{sec:certified-region}

We formalize the semi-strong setting as follows.
For each orientation, one fixed player is designated as the \emph{optimal agent} (AI) and always plays the canonical optimal move $m^{\star}(p)$ with respect to $V(\cdot)$.
The other player is a \emph{free agent} that may choose any legal move (e.g., a human).

Let $p_0$ denote the initial position.
Since either player may be designated as the optimal agent, we define two orientation-specific certified regions.

\paragraph{Orientation-specific certified regions.}
Let $R_{\text{first}}$ (resp.\ $R_{\text{second}}$) denote the set of positions that can occur in a play in which the first (resp.\ second) player is the optimal agent and always plays $m^{\star}(\cdot)$, while the opponent may play arbitrarily.
Equivalently, for $P\in\{\text{first},\text{second}\}$, $R_{P}$ is the smallest set satisfying:
\begin{enumerate}
\item $p_0\in R_{P}$.
\item If $p\in R_{P}$ and it is the \emph{free agent}'s turn at $p$, then for all $m\in M(p)$, the successor $p\cdot m \in R_{P}$.
\item If $p\in R_{P}$ is non-terminal and it is the \emph{optimal agent}'s turn at $p$, then the canonical-optimal successor
\[
p\cdot m^{\star}(p)\in R_{P}.
\]
\end{enumerate}

Intuitively, $R_{P}$ contains all positions that the free agent can force to occur, assuming the optimal agent responds with the fixed canonical optimal move whenever it is its turn.
A semi-strong solution for orientation $P$ aims to provide correct values for all positions in $R_{P}$ and the canonical optimal move at every non-terminal position in $R_{P}$ where the optimal agent is to move.

\paragraph{Semi-strong union region (orientation-agnostic certification).}
Our goal in this paper is to support a single artifact that works regardless of whether the optimal agent plays first or second.
Accordingly, we define the (orientation-agnostic) \emph{certified region} as the union
\[
R \;:=\; R_{\text{first}} \cup R_{\text{second}}.
\]
Unless stated otherwise, $R$ refers to this union region throughout the paper.
When an argument depends on a specific orientation, we will explicitly write $R_{\text{first}}$ or $R_{\text{second}}$.

\begin{definition}[Semi-strongly solved]
Fix the terminal utility $\texttt{value}(\cdot)$ and the tie-breaking rule $\mathrm{TB}$ (hence the canonical move $m^{\star}(\cdot)$).
The game is \textbf{semi-strongly solved} if, for each orientation $P\in\{\text{first},\text{second}\}$,
(i) the game-theoretic value $V(p)$ is determined for every position $p\in R_P$, and
(ii) for every non-terminal position $p\in R_P$ at which the designated optimal agent is to move, the canonical optimal move $m^{\star}(p)$ is determined.
Equivalently, semi-strong solving determines exact values on $R$ and the canonical move at every certified optimal-agent decision point.
\end{definition}

\subsection{Node Kinds and Certification Obligations}
\label{sec:obligations}

Our algorithm labels each visited search node by a \emph{node kind}
\[
k\in\{P, A', P', C, A\},
\]
which determines the \emph{certification obligation} to be satisfied at that node.

\paragraph{Informal meaning of node kinds.}
Kinds $P$, $A'$, and $P'$ correspond to nodes whose values and/or canonical move choices are required to certify correctness on the region $R$.
Kinds $C$ and $A$ correspond to auxiliary cut/all reasoning (fail-high / fail-low) in the sense of Knuth's classification of alpha--beta nodes \cite{knuth1975analysis-alphabeta}.
The kinds are generated by the transition rules (Algorithm~\ref{tab:alg1}) together with PV promotion in PVS/negascout search~\cite{pearl1980scout,pearl1980asymptotic,reinefeld1983negascout} (Algorithm~\ref{tab:alg2}).

\paragraph{Obligations as recursive requirements.}
Formally, each kind $k$ defines an obligation $\mathcal{O}_k$.
An obligation specifies not only what the current call must return (exact value, best move, or pruning bound), but also what obligations must hold for the children that are explored.
We will rely on the following obligation intuitions:

\begin{itemize}
\item $\mathcal{O}_P$ (\emph{PV-capable; side-to-move may be free or optimal}):
the node must be solved exactly under a full window, and a canonical optimal move must be identifiable.
Moreover, since the side to move at a $P$-node may be the free agent (depending on orientation and play history), \emph{all legal moves must be covered} at this node kind (i.e., beta-cutoff is not permitted as it could skip a move the free agent might choose).
The eventual PV child is required to satisfy the $P$ obligation, while every non-PV child is required to satisfy the $A'$ obligation.
\item $\mathcal{O}_{A'}$ (\emph{optimal agent to move}):
the node must identify the canonical optimal move and its exact value (full-window obligation).
The eventual PV child is required to satisfy the $P'$ obligation.
Non-PV siblings need only be refuted as required by alpha--beta pruning.
\item $\mathcal{O}_{P'}$ (\emph{free agent to move}):
all legal moves must be covered (no cutoff) in the sense that the returned value must remain correct under arbitrary choice by the free agent,
and all children are required to satisfy the $A'$ obligation.
\item $\mathcal{O}_{C}$ and $\mathcal{O}_{A}$ (\emph{auxiliary cut/all}):
only sound pruning bounds are required; any sound transposition-table information (exact value or bound) is sufficient for reuse.
\end{itemize}

\paragraph{PV promotion and obligation downgrades.}
When move ordering is imperfect, a non-first child may become the eventual PV child by raising $\alpha$, and is then re-searched under the PV obligation.
At the same time, a previously explored PV candidate is downgraded to a non-PV role.
Such downgrades are always safe: information obtained under a stronger obligation satisfies the requirements of any weaker later role.
For example, a child previously searched under the $P$ obligation remains valid when later treated under the $A'$ obligation, and any entry with an exact value suffices for later use as a $C$-node bound.
(We make this inclusion relation explicit when discussing transposition-table reuse in Section~\ref{sec:transposition-table}.)

\subsection{Outputs: Solution Artifact and Proof Certificate}
\label{sec:artifacts}

Our framework produces two conceptually distinct outputs.

\paragraph{Solution artifact.}
A \emph{solution artifact} is a persistent representation that supports (i) exact value queries for positions in the certified region $R$ and (ii) extraction of a canonical optimal move whenever required by the obligation model (in particular at $P$/$A'$ kinds).
Operationally, this is naturally realized as a dumped transposition table~\cite{transpositiontable1967} or an equivalent key--value store.
For deployment, the artifact only needs to store information sufficient to answer queries on $R$; it need not include auxiliary cut/all nodes explicitly. It may retain additional certified positions, so its row count need not equal the size of $R$.

\paragraph{Proof certificate.}
A \emph{proof certificate} is additional information sufficient for third-party verification of the solution artifact.
A proof certificate may be implemented as (a) a reference verifier that recomputes the necessary auxiliary reasoning (including revisiting $C$/$A$ nodes), (b) a fuller dump of transposition-table entries including auxiliary nodes, or (c) other proof logs.
Thus, unlike the solution artifact, a proof certificate is about \emph{verifiability} rather than \emph{deployability}~\cite{hurd2005verification}.

\paragraph{Remark on terminology.}
We use ``proof certificate'' to emphasize third-party verifiability of the computed result.
In some communities, similar objects are referred to as \emph{certificates} or \emph{artifacts}; we will consistently use the above terms throughout.

\section{Methods}

This section describes the algorithmic framework used to compute a semi-strong solution artifact and (optionally) a proof certificate.
All foundational definitions---the game model and value function $V(\cdot)$, the certified region $R$ (including orientation-specific regions), the node kinds and obligations, and the artifact/certificate terminology---are fixed in Section~\ref{sec:preliminaries}.
Here we focus on how to \emph{compute} and \emph{export} the certified information efficiently.

\subsection{Relationship to Existing Notions}
\label{sec:relationship}

We recall two standard reference points.
A \emph{strong solution} determines the game-theoretic value for all positions reachable under arbitrary play from the initial position, whereas a \emph{weak solution} determines the value of the initial position together with a strategy that guarantees at least that value from the start \cite{allis1994searching-thesis}.
Our semi-strong notion (Definition in Section~\ref{sec:certified-region}) lies between these two.

\begin{proposition}[Relationship to existing notions]
Strongly solved $\Rightarrow$ semi-strongly solved $\Rightarrow$ weakly solved.
\end{proposition}
\begin{proof}
A strong solution determines $V(p)$ for all reachable positions under arbitrary play, hence also for every $p\in R$.
Moreover, if $p\in R_P$ is non-terminal and the designated optimal agent is to move at $p$, then every legal successor $p\cdot m$ is reachable in the original game under arbitrary play.
A strong solution therefore determines $V(p\cdot m)$ for all $m\in M(p)$, and the fixed tie-breaking rule $\mathrm{TB}$ then determines the canonical optimal move $m^{\star}(p)$.
Hence the game is semi-strongly solved.

Conversely, assume the game is semi-strongly solved.
If $p_0$ is terminal, the weak-solution claim is immediate.
Otherwise, consider the orientation in which the first player is the optimal agent, and define a strategy $\sigma$ on non-terminal positions $p\in R_{\text{first}}$ with the first player to move by
\[
\sigma(p):=m^{\star}(p).
\]
By clauses~(1)--(3) in the definition of $R_{\text{first}}$, every play from $p_0$ in which the first player follows $\sigma$ and the second player moves arbitrarily remains in $R_{\text{first}}$.
By Proposition~\ref{prop:derived-game-view}, the value component on $R_{\text{first}}$ agrees with the derived game $G_{\text{first}}$, in which $\sigma$ is exactly the legal policy forced on the optimal agent.
Therefore $\sigma$ guarantees a first-player payoff of at least $V(p_0)$ against arbitrary second-player replies, with equality against optimal opposition, so the game is weakly solved.
\end{proof}

\paragraph{A Derived-Game View.}
The value component of semi-strong solving is a strong solution of an action-restricted game: restrict the optimal agent to its canonical move while retaining every free-agent move.

\begin{definition}[Derived game under a one-sided optimality constraint]
\label{def:derived-game}
Fix an orientation $P\in\{\text{first},\text{second}\}$ indicating which player is the optimal agent, and fix the canonical move function $m^{\star}(\cdot)$ (Section~\ref{sec:game-model}).
Define the derived game $G_P$ by keeping the same positions, terminal utility $\texttt{value}(\cdot)$, and turn structure as the original game, but modifying the legal-move rule as follows:
\begin{itemize}
\item If it is the free agent's turn at $p$, then the legal moves are $M(p)$ (unchanged).
\item If $p$ is non-terminal and it is the optimal agent's turn at $p$, then the legal moves are restricted to the singleton set $\{m^{\star}(p)\}$. Terminal positions retain an empty legal-move set.
\end{itemize}
\end{definition}

\begin{proposition}[Value component of semi-strong certification as a strong solution of a derived game]
\label{prop:derived-game-view}
For a fixed orientation $P$, the following are equivalent:
\begin{enumerate}
\item determining $V(p)$ for every position $p\in R_P$ (i.e., the value component of semi-strong solving for orientation $P$), and
\item strongly solving the derived game $G_P$ (Definition~\ref{def:derived-game}), i.e., determining the game-theoretic value for all positions reachable from $p_0$ in $G_P$.
\end{enumerate}
\end{proposition}
\begin{proof}[Proof sketch; full proof in Appendix~A.1]
By construction, the positions reachable from $p_0$ in $G_P$ are exactly those reachable when the optimal agent is forced to play $m^{\star}(\cdot)$ while the free agent may play any legal move; this reachable set is precisely $R_P$.
Moreover, the backed-up value recursion in $G_P$ agrees with the original $V(\cdot)$ on this reachable set (the only change is that max-choices at optimal-agent turns are replaced by the fixed maximizing move $m^{\star}(p)$).
Thus, solving $G_P$ on its reachable positions is exactly correctness of $V(\cdot)$ on $R_P$.
\end{proof}

\paragraph{Remark.}
Proposition~\ref{prop:derived-game-view} concerns the \emph{value component} of semi-strong solving.
The additional requirement that $m^{\star}(p)$ be determined at certified optimal-agent turns is what supports strategy extraction and a deployable solution artifact.

The union \(R=R_{\text{first}}\cup R_{\text{second}}\) supports both player orientations in a single artifact. The following proposition makes its weak-solution guarantee explicit.

\begin{proposition}[A semi-strong solution artifact yields a weak solution]
\label{prop:artifact-implies-weak}
Let $\mathcal{A}$ be a solution artifact for the certified region $R$ (Section~\ref{sec:certified-region} and Section~\ref{sec:artifacts}).
That is, $\mathcal{A}$ supports exact value queries $V(p)$ for every $p\in R$, and for each orientation $P\in\{\text{first},\text{second}\}$ it can return the canonical move $m^{\star}(p)$ at every non-terminal position $p\in R_P$ where the designated optimal agent is to move, consistently with the fixed tie-breaking rule $\mathrm{TB}$ (Section~\ref{sec:game-model}).
Then $\mathcal{A}$ determines (i) the game-theoretic value $V(p_0)$ of the initial position and (ii) a strategy for the first player from $p_0$ that guarantees a payoff of at least $V(p_0)$ against arbitrary replies.
In particular, $\mathcal{A}$ constitutes a weak solution in the usual sense \cite{allis1994searching-thesis}.
\end{proposition}

\begin{proof}[Proof Sketch; Full Proof in Appendix~A.2]
The artifact supplies \(V(p_0)\). In the first-player orientation, its canonical policy keeps every play inside \(R_{\text{first}}\) against arbitrary replies. By Proposition~\ref{prop:derived-game-view}, that policy guarantees at least \(V(p_0)\) to the first player; an optimal opponent can hold the payoff to \(V(p_0)\).
\end{proof}

\paragraph{Remark on ``reasonable resources.''}
Following community usage, one sometimes qualifies solving notions by ``under reasonable resources'' \cite{allis1994searching-thesis}.
We keep the \emph{solution notion} (correctness on $R$, together with canonical moves at certified optimal-agent turns) resource-independent, and instead specify resources operationally through the produced artifact format and the reported computational environment, position counts, and released artifacts.

\subsection{Reopening Alpha--Beta Algorithm}

We now describe an algorithmic framework to compute a semi-strong solution artifact for the certified region $R$.
The key idea is to switch the search obligation depending on the node kind (Section~\ref{sec:obligations}): some nodes must identify a canonical optimal move and therefore require a reopened (full) window, while auxiliary nodes can be processed using standard alpha--beta cut/all reasoning \cite{knuth1975analysis-alphabeta}.

\paragraph{Remarks on simplifying assumptions (analysis only).}
Our closed-form node-count analysis adopts standard idealizations from the alpha--beta literature \cite{knuth1975analysis-alphabeta}:
(i) no transpositions and no repetitions, (ii) optimal move ordering (the principal child is searched first), and (iii) unique backed-up values (no ties), which simplifies PV-versus-non-PV classification.
In practical games, ties may occur; our framework handles them via the fixed tie-breaking rule $\mathrm{TB}$ and the canonical move $m^{\star}(\cdot)$ (Section~\ref{sec:game-model}), without changing $V(\cdot)$.

\subsubsection{Kinds and Generation Rules of Nodes}
\label{section:generation-rule}

In reopening alpha--beta, each visited node is labeled by a \emph{kind} $k\in\{P, A', P', C, A\}$, which specifies the \emph{search obligation} (Section~\ref{sec:obligations}).
Briefly, $P$- and $A'$-nodes are \emph{agent-decision} nodes where we must be able to recover the canonical optimal move, so we reopen the search window to obtain an exact value and a certified best move.
$P'$-nodes are \emph{free-agent decision} nodes where the free agent may choose any move; we must cover all legal moves (no cutoff).
$C/A$-nodes are auxiliary cut/all nodes outside the certification requirements and are processed as standard alpha--beta nodes.

The node kinds and their generation rules are as follows.
Under the optimal-ordering assumption, the first child is the principal child.

\begin{enumerate}
\item \textbf{P-node:} The initial position is a $P$-node. The principal child of a $P$-node is a $P$-node, while all non-principal children are $A'$-nodes. $P$ corresponds to Knuth's Type-1 (PV) node \cite{knuth1975analysis-alphabeta}.
\item \textbf{$A'$-node:} The principal child of an $A'$-node is a $P'$-node, and all non-principal children are $C$-nodes.
\item \textbf{$P'$-node:} All children of a $P'$-node are $A'$-nodes.
\item \textbf{C-node:} Each $C$-node has only one visited child (due to beta-cutoff / fail-high), which is an $A$-node. $C$ corresponds to Knuth's Type-2 (cut / fail-high) node \cite{knuth1975analysis-alphabeta}.
\item \textbf{A-node:} All children of an $A$-node are $C$-nodes. $A$ corresponds to Knuth's Type-3 (all / fail-low) node \cite{knuth1975analysis-alphabeta}.
\end{enumerate}

\begin{figure}[htbp]
\centering
\includegraphics[width=1.0\linewidth]{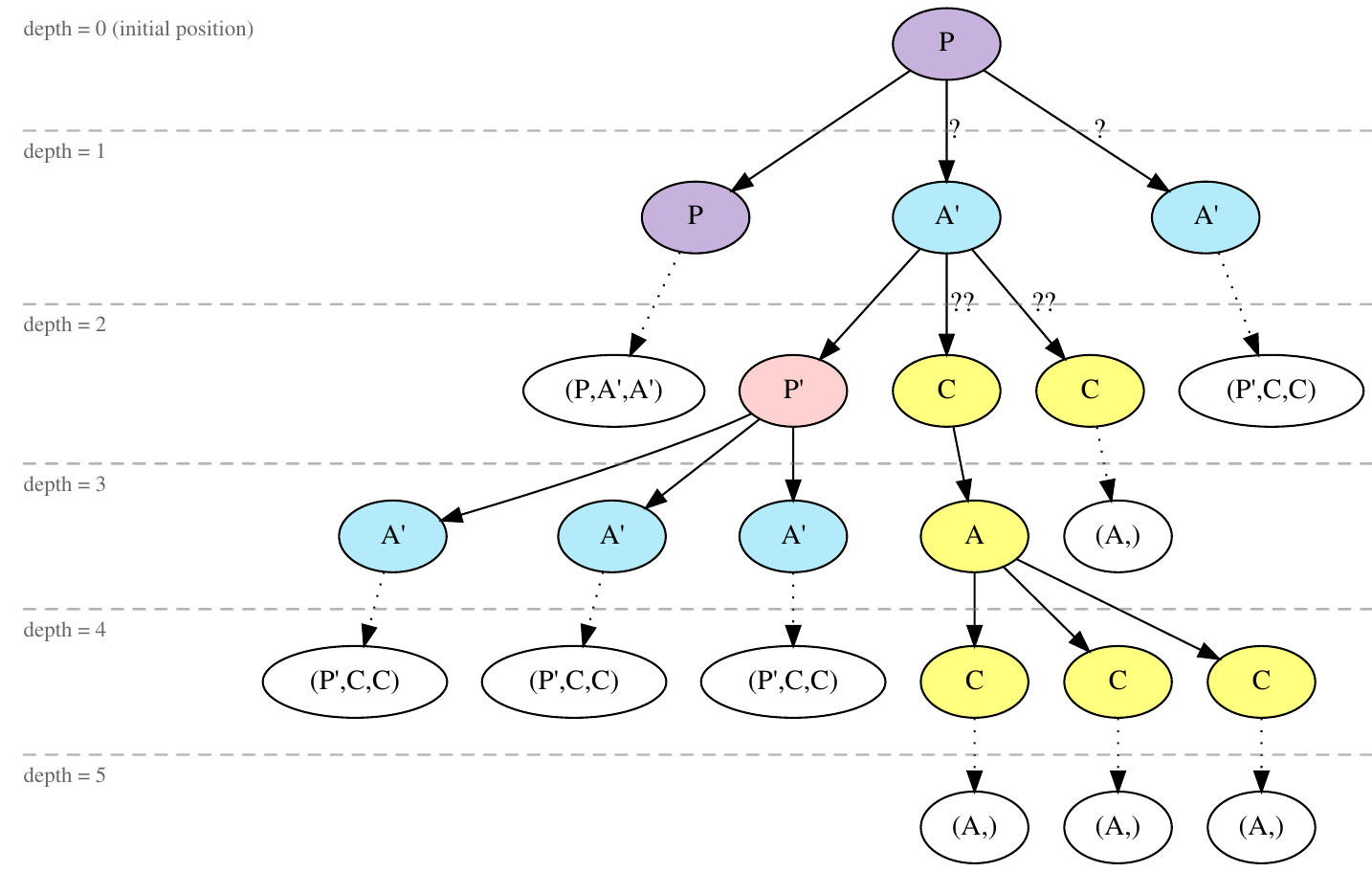}
\Description{Schematic diagram of node kinds in the reopening alpha-beta example.}
\caption{
Schematic node-kind structure for branching factor $b=3$ under the idealized assumptions of Section~3.2.1.
Colors indicate node kinds and their associated obligations: $P$-nodes (purple) and $A'$-nodes (blue) require identifying the canonical optimal move together with the exact value; $P'$-nodes (red) are free-agent choice points for which all legal moves must be covered; and $C$/$A$-nodes (yellow) are auxiliary cut/all nodes.
Horizontal guide lines mark depth levels, with depth~0 corresponding to the initial position.
Edges labeled ``?'' or ``??'' denote non-principal moves under optimal ordering.
White ellipses at the ends of dotted edges denote omitted subtrees; the label in each ellipse lists the kinds of the omitted node's children as an ordered tuple (principal child first), with singleton tuples written using a trailing comma, e.g., $(A,)$.
}
\label{fig:fig-reopening-ab-example}
\end{figure}

\subsubsection{Intuition for Node Kinds (Certified and Auxiliary)}

Along the canonical principal variation, $P$-nodes support both choices of optimal agent and therefore require exact values and canonical optimal moves. The first departure reaches an $A'$-node and fixes the deviating player as the free agent. This departure may be value-worsening or a different tied optimal move.

At an $A'$-node, reopening certifies the exact local value and the optimal agent's canonical response. Under ideal move ordering, the canonical child is searched first with the $P'$ obligation, yielding the value $v$. Each other move requires only an ordinary alpha--beta null-window test showing that its value, from the parent's perspective, is at most $v$. Equal-valued alternatives are resolved by the fixed tie-breaking rule. Such a test starts with kind $C$, and its descendants remain within the $C/A$ kinds; no reopening is required within that auxiliary search. With fixed branching factor $b$, ideal ordering, and no transpositions, an auxiliary subtree of remaining move depth $h$ therefore requires $O(b^{h/2})$ visited nodes, as in classical alpha--beta.

The canonical continuation retains the $P'$ obligation: every legal free-agent move is covered, and each resulting child is again an $A'$-node. Thus the same human may make value-worsening moves repeatedly, with reopening certifying each subsequent AI decision. The overall $O(d\,b^{d/2})$ bound includes these recurring certification obligations and the auxiliary searches attached throughout the region.

\subsubsection{Complexity Analysis}
\label{sec:complexity}

We use a uniform tree with fixed branching factor $b\ge2$. For the recurrences below, levels are numbered from $d=1$ at the root to $d=D$ at the leaves; thus level $d$ is move depth $d-1$ in Figure~\ref{fig:fig-reopening-ab-example}, and the maximum move depth is $D-1$. This one-level shift does not change the stated asymptotic bounds for fixed $b$. The number of nodes of kind $Y\in\{P,A',P',C,A\}$ at level $d$ is denoted by $N(Y,d)$.

\begin{itemize}
\item \textbf{P-node:} For any $d\in [1,D]$, we have $N(P,d)=1$.
\item \textbf{A'-node:} $N(A',1)=0$, and for $d>1$,
$N(A',d)=(b-1)N(P,d-1)+bN(P',d-1)$.
\item \textbf{P'-node:} $N(P',1)=0$, and for $d>1$,
$N(P',d)=N(A',d-1)$.
\item \textbf{C-node:} $N(C,1)=0$, and for $d>1$,
$N(C,d)=(b-1)N(A',d-1)+bN(A,d-1)$.
\item \textbf{A-node:} $N(A,1)=0$, and for $d>1$,
$N(A,d)=N(C,d-1)$.
\end{itemize}

We now derive closed forms. The key step is that $A'$-nodes satisfy a simple second-order recurrence, which then induces one for $C$-nodes.

\begin{lemma}[Closed form for $A'$- and $P'$-nodes]
\label{lem:AprimePprime}
For all $d\ge 1$,
\[
N(A',d)=b^{\left\lceil\frac{d-1}{2}\right\rceil}-1,
\qquad
N(P',d)=b^{\left\lceil\frac{d-2}{2}\right\rceil}-1.
\]
\end{lemma}

\begin{proof}[Proof sketch; full proof in Appendix~A.3]
Using $N(P,d)=1$ and $N(P',d)=N(A',d-1)$ for $d>1$, the recurrence for $A'$ becomes, for $d>2$,
\[
N(A',d)=(b-1)+b\,N(A',d-2),
\]
with bases $N(A',1)=0$ and $N(A',2)=b-1$.
Solving separately on even/odd $d$ yields $N(A',2k)=b^k-1$ and $N(A',2k+1)=b^k-1$, which is equivalently
$N(A',d)=b^{\lceil(d-1)/2\rceil}-1$.
Finally, $N(P',d)=N(A',d-1)$ gives the stated form for $P'$.
\end{proof}

\begin{lemma}[Closed form for $C$- and $A$-nodes]
\label{lem:CA}
For all $d\ge 1$,
\[
N(C,d)=\left\lceil \frac{d-2}{2} \right\rceil b^{\left\lceil \frac{d}{2} \right\rceil}
-\left\lceil \frac{d}{2} \right\rceil b^{\left\lceil \frac{d-2}{2} \right\rceil}+1,
\]
and
\[
N(A,d)=\left\lceil \frac{d-3}{2} \right\rceil b^{\left\lceil \frac{d-1}{2} \right\rceil}
-\left\lceil \frac{d-1}{2} \right\rceil b^{\left\lceil \frac{d-3}{2} \right\rceil}+1.
\]
\end{lemma}

\begin{proof}[Proof sketch; full proof in Appendix~A.4]
From $N(A,d)=N(C,d-1)$ (for $d>1$) and the recurrence for $C$, we obtain for $d>2$
\[
N(C,d)=(b-1)N(A',d-1)+b\,N(C,d-2).
\]
By Lemma~\ref{lem:AprimePprime}, $N(A',d-1)=b^{\lceil(d-2)/2\rceil}-1$.
Solving the resulting second-order linear recurrence by splitting into even/odd levels and unrolling the geometric sums yields the stated closed form for $N(C,d)$.
Finally, $N(A,d)=N(C,d-1)$ gives the expression for $N(A,d)$.
\end{proof}

\begin{corollary}[Overall node count and asymptotic complexity]
\label{cor:overall-complexity}
Let
\[
T(D)=\sum_{d=1}^{D}\sum_{Y\in\{P,A',P',C,A\}} N(Y,d)
\]
be the total number of visited nodes in the first $D$ levels.
Then $T(D)=O(D\,b^{D/2})$.
\end{corollary}

\begin{proof}
For each fixed level $d$, we bound the contribution of each node kind.
First, $N(P,d)=1=O\!\left(b^{\lceil d/2\rceil}\right)$ for every $d\ge 1$.
By Lemma~\ref{lem:AprimePprime},
\[
N(A',d)=b^{\left\lceil\frac{d-1}{2}\right\rceil}-1,
\qquad
N(P',d)=b^{\left\lceil\frac{d-2}{2}\right\rceil}-1,
\]
so both $N(A',d)$ and $N(P',d)$ are $O\!\left(b^{\lceil d/2\rceil}\right)$.
By Lemma~\ref{lem:CA}, the closed forms for $N(C,d)$ and $N(A,d)$ are each a linear combination of two terms of the form
\[
\left\lceil \frac{d-c_1}{2} \right\rceil b^{\left\lceil \frac{d-c_2}{2} \right\rceil}
\]
plus a constant, and therefore
\[
N(C,d)=O\!\left(d\,b^{\lceil d/2\rceil}\right),
\qquad
N(A,d)=O\!\left(d\,b^{\lceil d/2\rceil}\right).
\]
Hence
\[
\sum_{Y\in\{P,A',P',C,A\}} N(Y,d)
= O\!\left(d\,b^{\lceil d/2\rceil}\right).
\]
Summing over levels gives
\[
T(D)
=\sum_{d=1}^{D} O\!\left(d\,b^{\lceil d/2\rceil}\right).
\]
Since $d\le D$ for $1\le d\le D$,
\[
\sum_{d=1}^{D} d\,b^{\lceil d/2\rceil}
\le D\sum_{d=1}^{D} b^{\lceil d/2\rceil}
\le 2D\sum_{j=1}^{\lceil D/2\rceil} b^{j}
= O\!\left(D\,b^{D/2}\right)
\qquad (b\ge 2).
\]
Therefore
\[
T(D)=O(D\,b^{D/2}).
\]
The degenerate case $b=1$ is immediate.
\end{proof}

\paragraph{Relation to Knuth's minimal alpha--beta tree.}
Under the same idealized assumptions, standard alpha--beta visits exactly the critical positions and the resulting minimal alpha--beta tree has size $\Theta(b^{d/2})$ \cite{knuth1975analysis-alphabeta}.
Reopening alpha--beta introduces two additional certified node classes, $A'$ and $P'$, which arise from the semi-strong requirement to identify canonical best moves not only on the principal variation ($P$-nodes) but also after a free-agent deviation ($A'$-nodes), while allowing arbitrary free-agent moves at $P'$-nodes.
This yields $O(d\,b^{d/2})$ visited nodes under the stated idealizations. The extra depth factor accounts for auxiliary $C/A$ refutations attached throughout the certified region. By Lemma~\ref{lem:AprimePprime}, the union of the certified $P$, $A'$, and $P'$ nodes alone has $\Theta(b^{d/2})$ nodes in this fixed-$b$ tree model; it should not be confused with the larger searched proof structure.

\subsection{Algorithm Description}
\label{sec:algorithm-description}

Algorithm~\ref{tab:alg2} is a negamax alpha--beta/PVS search whose behavior is controlled by the node kind (Section~\ref{sec:obligations}).
We use the term \emph{principal variation (PV)} for the sequence of moves induced by the canonical principal-child choice; in the presence of ties, PV selection is made canonical by the global tie-breaking rule $\mathrm{TB}$ (Section~\ref{sec:game-model}).

\paragraph{Determinism and Canonical Move Selection.}
For Algorithm~\ref{tab:alg2}, fix a deterministic move order as a function of the position. Among moves with equal backed-up value, \(\mathrm{TB}\) selects the earliest in this order. A later move is promoted to PV only after a strict score improvement. If an implementation changes its search order using history or TT information, it must keep the canonical tie-breaking rule separate or freeze a consistent selected policy; deterministic code alone does not imply a history-independent choice.

\paragraph{Reopening and Cutoffs.}
As printed, Algorithm~\ref{tab:alg2} reopens the window at all three certified kinds \(P,A',P'\). At \(P\) and \(A'\), it also identifies the canonical move. At \(P\) and \(P'\), it must cover all legal moves, so cutoff is disabled. Auxiliary \(C/A\) calls use ordinary alpha--beta pruning.

\paragraph{Finite Full Windows.}
We write the full window as \([-\infty,+\infty]\). A finite implementation must use endpoints strictly outside the attainable score range: if integer values satisfy \(|V(p)|\le U\), \([-U-1,U+1]\) suffices. Simply replacing infinity by \(U\) in Algorithm~\ref{tab:alg2} is unsafe: a later move scoring exactly \(U\) would fail the strict promotion test on line~17 and could leave its required PV obligation uncertified. An upper-bound shortcut is sound only after the canonical principal child and all other required coverage have been certified.

\paragraph{Null-window notation for integer-valued games.}
Throughout this paper, terminal utilities and backed-up values are integer-valued.
Accordingly, in null-window calls we take $\epsilon=1$, so windows such as $[-\alpha-\epsilon,-\alpha]$ and $[\beta-\epsilon,\beta]$ are the usual one-unit null windows.

\begin{algorithm}[htbp]
\caption{\texttt{GetChildNodeKind} returns the child kind. The second argument indicates whether the child is on the (eventual) principal variation.}
\label{tab:alg1}
\hbox to\hsize{\hfil
\begin{tabular}{r|l}\hline
name: & GetChildNodeKind \\\hline
input: & $k$: string, is\_principal\_child: boolean \\\hline
output: & string \\\hline
1: & if $k=$ "P":\\\hline
2: & $\quad$return "P" if is\_principal\_child else "A'"\\\hline
3: & if $k=$ "A'":\\\hline
4: & $\quad$return "P'" if is\_principal\_child else "C"\\\hline
5: & if $k=$ "P'":\\\hline
6: & $\quad$return "A'"\\\hline
7: & if $k=$ "C":\\\hline
8: & $\quad$return "A"\\\hline
9: & if $k=$ "A":\\\hline
10: & $\quad$return "C"\\\hline
11: & assert False\\\hline
\end{tabular}\hfil}
\end{algorithm}

\begin{algorithm}[htbp]
\caption{Reopening alpha--beta with PVS/negascout-style re-search (fail-soft~~\cite{failsoft-alphabeta-1983}). Move ordering is deterministic; the exploration order at each position defines $\mathrm{TB}$ among equal-scoring moves.}
\label{tab:alg2}
\hbox to\hsize{\hfil
\begin{tabular}{r|l}\hline
name: & Search \\\hline
input: & $p$: position, $\alpha$: scalar, $\beta$: scalar, $k$: string \\\hline
output: & scalar \\\hline
1: & if $p$ is terminal: \\\hline
2: & $\quad$return value($p$) \\\hline
3: & if $k\in \{"P","A'","P'"\}$: \\\hline
4: & $\quad\alpha\leftarrow -\infty$ \\\hline
5: & $\quad\beta\leftarrow +\infty$ \\\hline
6: & $s\leftarrow -\infty$ \\\hline
7: & $M\leftarrow$ all legal moves in $p$ \\\hline
8: & Sort $M$ deterministically in descending order of promise \\\hline
9: & for each $m\in M$ in order: \\\hline
10: & $\quad p'\leftarrow$ the position after applying $m$ to $p$ \\\hline
11: & $\quad$if $m$ is the first move in the loop: \\\hline
12: & $\quad\quad k'\leftarrow$ GetChildNodeKind($k$, True) \\\hline
13: & $\quad\quad$score $\leftarrow$ -Search($p'$, -$\beta$, -$\alpha$, $k'$) \\\hline
14: & $\quad$else: \\\hline
15: & $\quad\quad k_{\text{nonPV}}\leftarrow$ GetChildNodeKind($k$, False) \\\hline
16: & $\quad\quad$score $\leftarrow$ -Search($p'$, -$\alpha-\epsilon$, -$\alpha$, $k_{\text{nonPV}}$) \\\hline
17: & $\quad\quad$if $\alpha<$ score $<\beta$: \\\hline
18: & $\quad\quad\quad k_{\text{PV}}\leftarrow$ GetChildNodeKind($k$, True) \\\hline
19: & $\quad\quad\quad$score $\leftarrow$ -Search($p'$, -$\beta$, -$\alpha$, $k_{\text{PV}}$) \\\hline
20: & $\quad\alpha\leftarrow \max(\alpha,$ score$)$ \\\hline
21: & $\quad s\leftarrow \max(s,$ score$)$ \\\hline
22: & $\quad$if $k \notin \{"P","P'"\}$ and $\alpha\ge \beta$: \\\hline
23: & $\quad\quad$return $\alpha$ \hfill \textit{// beta-cutoff} \\\hline
24: & return $s$ \hfill \textit{// fail-soft} \\\hline
\end{tabular}\hfil}
\end{algorithm}

\paragraph{Principal-child semantics vs.\ move ordering.}
In Algorithm~\ref{tab:alg1}, the boolean argument of \texttt{GetChildNodeKind} denotes whether the child lies on the \emph{eventual} principal variation (PV), not merely whether it is first in the initial move ordering.
When move ordering is imperfect, a non-first child may become the PV child by \emph{strictly} raising $\alpha$; Algorithm~\ref{tab:alg2} handles this by re-searching such a child under the PV obligation (lines~17--19).
Equal-score later children are not promoted: by the convention above, the earliest explored maximizing child is already the canonical PV child.

\paragraph{PV promotion and obligation downgrades.}
When move ordering is imperfect, Algorithm~\ref{tab:alg2} may promote a non-first child to the eventual PV child via the re-search (lines~17--19), upgrading that child to the PV obligation.
At the same time, the previously explored PV candidate is downgraded to a non-PV role.
This downgrade is sound: information obtained under a stronger obligation remains valid when later treated under a weaker role (Section~\ref{sec:obligations}).

\paragraph{An obligation-oriented formulation.}
For exposition, we also provide a specification-level \emph{obligation-oriented} formulation corresponding to Algorithm~\ref{tab:alg2}, shown as Algorithm~\ref{tab:alg2-1}.
This version makes the obligation structure explicit by branching directly on the node kind and by selecting the canonical principal move $m^{\star}(p)$ (Section~\ref{sec:game-model}).

Algorithm~\ref{tab:alg2-1} is a specification of the canonical-first certification phase, rather than a procedure for finding \(m^{\star}(p)\). Its root call has a full window. The calls generated at certified kinds inherit full windows; only auxiliary refutations use narrow windows. At an \(A'\)-node, a known payoff upper bound can justify omitting the remaining refutations once the canonical principal child is fully certified. At \(P\) and \(P'\), all legal moves must still be covered.

\begin{algorithm}[htbp]
\caption{A specification-level obligation-oriented formulation corresponding to Algorithm~\ref{tab:alg2}.}
\label{tab:alg2-1}
\hbox to\hsize{\hfil
\begin{tabular}{r|l}\hline
name: & Search\_Spec \\\hline
input: & $p$: position, $\alpha$: scalar, $\beta$: scalar, $k$: string \\\hline
output: & scalar \\\hline
1:  & if $p$ is terminal: \\\hline
2:  & \quad return value($p$) \\\hline
3:  & $s\leftarrow -\infty$ \\\hline
4:  & $M\leftarrow$ all legal moves in $p$ \\\hline
5:  & $\lambda(m)\leftarrow$ the position after applying move $m$ to $p$ \\\hline
6:  & $m^{\star}\leftarrow m^{\star}(p)$ \hfill \textit{// canonical optimal move; Section~\ref{sec:game-model}} \\\hline
7:  & if $k\in\{"P","P'"\}$: \\\hline
8:  & \quad $s\leftarrow$ -Search\_Spec($\lambda(m^{\star})$, -$\infty$, $\infty$, GetChildNodeKind($k$, True)) \\\hline
9:  & \quad for each $m\in M\setminus\{m^{\star}\}$ in descending order of promise: \\\hline
10: & \quad\quad $s'\leftarrow$ -Search\_Spec($\lambda(m)$, -$\infty$, $\infty$, GetChildNodeKind($k$, False)) \\\hline
11: & \quad\quad assert $s'\leq s$ \hfill \textit{// non-principal moves are not better}\\\hline
12: & else if $k\in\{"A'","A"\}$: \\\hline
13: & \quad $s\leftarrow$ -Search\_Spec($\lambda(m^{\star})$, -$\beta$, -$\alpha$, GetChildNodeKind($k$, True)) \\\hline
14: & \quad if $s\geq \beta$: \\\hline
15: & \quad\quad return $s$ \hfill \textit{// beta-cutoff (fail-soft)}\\\hline
16: & \quad $\alpha\leftarrow \max(\alpha,s)$ \\\hline
17: & \quad for each $m\in M\setminus\{m^{\star}\}$ in descending order of promise: \\\hline
18: & \quad\quad $s'\leftarrow$ -Search\_Spec($\lambda(m)$, -$\alpha-\epsilon$, -$\alpha$, GetChildNodeKind($k$, False)) \\\hline
19: & \quad\quad assert $s'\leq \alpha$ \hfill \textit{// null-window refutation succeeds}\\\hline
20: & else: \hfill \textit{// $k="C"$} \\\hline
21: & \quad $s\leftarrow$ -Search\_Spec($\lambda(m^{\star})$, -$\beta$, -$\alpha$, GetChildNodeKind($k$, True)) \\\hline
22: & \quad assert $s\geq \beta$ \hfill \textit{// a C-node is, by definition, fail-high}\\\hline
23: & return $s$ \hfill \textit{// fail-soft}\\\hline
\end{tabular}\hfil}
\end{algorithm}

\paragraph{Correctness with respect to obligations.}
The next proposition states the main semantic guarantee of Algorithm~\ref{tab:alg2}.
It does not require the ideal move ordering, absence of ties, or absence of repeated states used for the closed-form tree counts; it does require the fixed move order and full windows specified above.

\begin{proposition}[Obligation correctness of reopening alpha--beta]
\label{prop:search-correctness}
Assume that recursive evaluation is well-founded (equivalently, every recursive call chain terminates).
With a fixed position-dependent exploration order and the full windows in Algorithm~\ref{tab:alg2}, every call of \texttt{Search} on input $(p,\alpha,\beta,k)$, where $\alpha<\beta$, satisfies the obligation $\mathcal O_k$ from Section~\ref{sec:obligations}.
More explicitly, with the child and move assertions below applying to non-terminal positions:
\begin{enumerate}
\item if $k\in\{\texttt{"P"},\texttt{"A'"},\texttt{"P'"}\}$, the returned score is the exact value $V(p)$;
\item if $k\in\{\texttt{"P"},\texttt{"A'"}\}$, the eventual PV child is the canonical optimal move $m^{\star}(p)$;
\item if $k=\texttt{"P"}$, the eventual PV child satisfies the $P$-obligation and every non-PV child satisfies the $A'$-obligation;
\item if $k=\texttt{"A'"}$, the eventual PV child satisfies the $P'$-obligation and every non-PV child is soundly refuted under ordinary alpha--beta reasoning;
\item if $k=\texttt{"P'"}$, every child satisfies the $A'$-obligation;
\item if $k\in\{\texttt{"C"},\texttt{"A"}\}$, the returned score is alpha--beta sound for the input window $[\alpha,\beta]$: an interior return is exact, a fail-high return is a valid lower bound, and a fail-low return is a valid upper bound.
\end{enumerate}
\end{proposition}

\begin{proof}
We argue by induction on the remaining play length from $p$.
The terminal case is immediate from lines~1--2.

Assume $p$ is non-terminal.
For $k\in\{\texttt{"P"},\texttt{"P'"}\}$, lines~3--5 reopen the window to $[-\infty,+\infty]$, and line~22 disables cutoff.
Hence every legal child is searched to completion.
By the induction hypothesis, each recursive child call at these kinds returns the exact child value under the obligation prescribed by \texttt{GetChildNodeKind}.
The loop therefore computes
\[
\max_{m\in M(p)} -V(p\cdot m)=V(p).
\]
Moreover, lines~17--19 promote a later child only when it \emph{strictly} improves the current score.
Therefore the eventual PV child is exactly the earliest maximizing move in the deterministic exploration order, i.e.\ the canonical optimal move $m^{\star}(p)$.
The child-obligation claims for $P$ and $P'$ follow directly from \texttt{GetChildNodeKind} together with the fact that a promoted child is re-searched under the PV obligation, whereas a demoted former PV child has already been solved under a stronger obligation than its later non-PV role requires.

For $k=\texttt{"A'"}$, lines~3--5 again reopen the window to a full range, so the current PV candidate is always solved exactly under the $P'$ obligation.
Each later child is first tested under the non-PV kind $C$.
If its score does not exceed the current $\alpha$, that non-PV refutation suffices.
If its score strictly exceeds $\alpha$, lines~17--19 re-search it under the PV obligation $P'$, after which it becomes the new PV candidate.
At loop end, the algorithm has therefore identified the exact maximum child value and the earliest maximizing child in the deterministic exploration order.
Thus the return value is $V(p)$ and the selected move is $m^{\star}(p)$, with the required $P'$ obligation on the eventual principal child.

Finally, for $k\in\{\texttt{"C"},\texttt{"A"}\}$, Algorithm~\ref{tab:alg2} is an ordinary fail-soft negamax alpha--beta search, differing only in the node-kind labels passed to recursive calls.
By the induction hypothesis, each recursive call returns information that is sound for its own window.
The usual alpha--beta invariant therefore applies: a return inside the input window is exact, a fail-high return is a valid lower bound, and a fail-low return is a valid upper bound.
This is exactly the requirement of $\mathcal O_C$ and $\mathcal O_A$.
\end{proof}

In particular, the root call \texttt{Search}$(p_0,-\infty,+\infty,\texttt{"P"})$ returns the exact initial value, and the information accumulated at $P$/$A'$/$P'$ nodes contains the data required for a semi-strong solution artifact. Imperfect ordering can also certify provisional PV branches outside the final $R$. Once canonical moves are fixed, a traversal of the defining rules for $R$ selects the required entries; additional certified entries may be retained or discarded.

\subsubsection{Correctness of the Node-Kind Generation Model}
\label{sec:node-kind-correctness}

In the complexity analysis (Section~\ref{sec:complexity}), we model the visited search tree using the five node kinds $\{P,A',P',C,A\}$ and the kind transition rules.
The following proposition justifies that this abstraction matches the behavior of Algorithm~\ref{tab:alg2} under the simplifying assumptions used for the closed-form node-count analysis.

\begin{proposition}[Node-kind correspondence under idealized assumptions]
\label{prop:node-kind-correspondence}
Assume (i) no transpositions and no repetitions, (ii) optimal move ordering (the principal child is always searched first), and (iii) distinct backed-up values (no ties).
Then, when Algorithm~\ref{tab:alg2} is invoked from the root with kind $k=\texttt{"P"}$, the induced visited search tree is exactly characterized by the node kinds $\{P,A',P',C,A\}$ and the generation rules in Section~\ref{section:generation-rule}:
each visited edge follows the kind transition produced by \texttt{GetChildNodeKind} (Algorithm~\ref{tab:alg1}), and the non-principal parts of the tree exhibit the standard cut/all structure \cite{knuth1975analysis-alphabeta}.
\end{proposition}

\begin{proof}[Proof sketch; full proof in Appendix~A.5]
Under assumption (ii), the first explored child in each loop is the eventual principal child, so the re-search mechanism (lines~17--19 of Algorithm~\ref{tab:alg2}) never triggers and the search follows the idealized ``principal child vs.\ non-principal siblings'' pattern.
At kinds $\texttt{"P"}$ and $\texttt{"A'"}$, Algorithm~\ref{tab:alg2} reopens the window to a full range (lines~3--5), enforcing the exact-value-and-best-move obligation.
At kinds $\texttt{"P"}$ and $\texttt{"P'"}$, beta-cutoff is disabled (line~22), ensuring that all legal moves are covered at free-agent choice points.
For other nodes, the search behaves as standard alpha--beta: with distinct values and optimal ordering, the non-principal siblings beyond the principal child become fail-high/fail-low cases that match Knuth's Type-2/Type-3 (cut/all) structure \cite{knuth1975analysis-alphabeta}, yielding the one-child alternation captured by the $C\leftrightarrow A$ transition.
Finally, Algorithm~\ref{tab:alg1} encodes exactly the kind transitions stated in Section~\ref{section:generation-rule}.
\end{proof}

\subsubsection{Transposition Table, Solution Artifact, and Proof Certificate}
\label{sec:transposition-table}

Algorithm~\ref{tab:alg2} is presented without a transposition table (TT)~\cite{transpositiontable1967} only to simplify the idealized analysis above.
In practical games (including Othello and Connect Four), transpositions are abundant and a TT is essential for efficiency.
Moreover, a TT (or an equivalent key--value store) provides a natural substrate for exporting the computed result as a solution artifact (Section~\ref{sec:artifacts}) and, if desired, as part of a proof certificate.

\paragraph{TT entry schema.}
We store TT entries keyed by positions (and, if relevant, side-to-move and rule-state such as repetition counters)~\cite{Zobrist1970TR}.
Each entry typically consists of:
(i) a depth (or remaining-depth) field,
(ii) a bound type (\textsc{Exact}, \textsc{Lower}, \textsc{Upper}),
(iii) the corresponding scalar value,
and optionally (iv) a best move to support direct canonical-move selection at kinds that require it.
This is standard in alpha--beta implementations; the only additional subtlety here is that in reopening alpha--beta, the \emph{node kind} determines the \emph{obligation strength} under which the information was produced.

\paragraph{Obligation inclusion for TT reuse.}
We treat each node kind $k\in\{P,A',P',C,A\}$ as defining a certification obligation $\mathcal{O}_k$ (Section~\ref{sec:obligations}).
A TT entry produced under $\mathcal{O}_k$ is reusable for a later query of kind $k'$ if it suffices to satisfy $\mathcal{O}_{k'}$, including recursively required child obligations (and PV promotion).
These obligations form a preorder (a partial order after identifying equivalent obligations): $\mathcal{O}_P$ is strongest; $\mathcal{O}_C$ and $\mathcal{O}_A$ are weakest; and $\mathcal{O}_{A'}$ and $\mathcal{O}_{P'}$ are generally incomparable.

\begin{proposition}[Sound TT reuse rules by node kind]
\label{prop:tt-reuse}
Assume the TT stores sound alpha--beta information at the stored depth (exact values and/or valid bounds).
Assume also that the canonical policy is unchanged and that recursively required certified information remains available. Then the following are sufficient reuse rules; they do not exclude additional reuse justified by inspecting or combining certificates:

\begin{enumerate}
\item If a position is queried as a $P$-node, the search can be omitted if it has already been stored under the $P$-node obligation.
\item If a position is queried as an $A'$-node, the search can be omitted if it has already been stored under the $P$-node or $A'$-node obligation.
\item If a position is queried as a $P'$-node, the search can be omitted if it has already been stored under the $P$-node or $P'$-node obligation.
\item If a position is queried as a $C$-node or an $A$-node, the search can be omitted (or TT bounds may be used) regardless of which node kind produced the entry, provided the bound type/depth is sufficient for the current window.
\end{enumerate}
\end{proposition}

\begin{proof}[Proof sketch; full proof in Appendix~A.6]
We compare obligations rather than only returned scalar bounds.

\emph{(i) $P$-nodes.}
$\mathcal{O}_P$ is strongest: it requires a full-window exact value, no cutoff, and the child-wise certification structure (principal child under $P$, non-principal children under $A'$), with possible PV promotions.
A TT package that already discharges $\mathcal{O}_P$, including the required descendants, therefore suffices to omit the search.

\emph{(ii) $A'$-nodes.}
$\mathcal{O}_{A'}$ requires an exact value and a canonical best move, and it requires that the eventual principal child satisfies the $P'$ obligation.
An entry produced under $\mathcal{O}_P$ implies this; an entry produced under $\mathcal{O}_{A'}$ is reusable by definition.
An entry produced under $\mathcal{O}_{P'}$ is not sufficient in general because $\mathcal{O}_{P'}$ does not enforce the specific $P'$-principal-child obligation required by $\mathcal{O}_{A'}$.

\emph{(iii) $P'$-nodes.}
$\mathcal{O}_{P'}$ requires that \emph{all} children satisfy the $A'$ obligation (free agent may choose any move), and no cutoff.
An entry produced under $\mathcal{O}_P$ implies this; an entry produced under $\mathcal{O}_{P'}$ is reusable by definition.
An entry produced under $\mathcal{O}_{A'}$ is not sufficient in general because $\mathcal{O}_{A'}$ only enforces a $P'$ obligation on the principal child and does not require that all children are solved as $A'$.

\emph{(iv) $C/A$-nodes.}
$C$- and $A$-nodes correspond to standard fail-high/fail-low reasoning where TT entries are used only as bounds for pruning, so any sound entry (exact value or valid bound at sufficient depth) can be used.
\end{proof}

\section{Experimental Results}

\paragraph{Computational environment.}
Unless otherwise noted, all experiments reported in this section were run on a single workstation
(AMD Ryzen~9~5950X, 128\,GB RAM) under Ubuntu~20.04, using up to 32 hardware threads.

\paragraph{Measurement Scope.}
The comparisons in this section use distinct-position counts for the primary search and certification structures and retained records, not runtime or compressed file size.
Othello counts exclude auxiliary move-selection searches; the Connect Four experiment assumes an existing exact-value oracle and excludes its construction cost.
The game-specific counting conventions are specified below, with further procedural details in Appendix~\hyperref[app:experimental-details]{C}.

\subsection{Semi-Strongly Solving \(6\times 6\) Othello}

We evaluate our framework by semi-strongly solving \(6\times 6\) Othello using reopening alpha--beta (Section~\ref{sec:algorithm-description}) and exporting a solution artifact for the certified region \(R\) (Section~\ref{sec:certified-region}).
The full rules are given in Appendix~\hyperref[app:rules-othello]{B.1}; the terminal utility used in our experiments is specified in Section~\ref{section:rule-othello}.

\subsubsection{Experimental Setup}
We chose \(6\times 6\) Othello for three reasons.
First, Othello has no repetition along a play: the game graph is acyclic (unlike chess), although it contains many transpositions.
Second, the terminal utility is naturally integer-valued (a score difference) rather than binary, so certifying exact values on \(R\) yields a quantitatively meaningful guarantee under arbitrary free-agent deviations.
Third, Othello is a widely studied and widely played benchmark game, which facilitates reproduction and comparison.
The reduced board size further makes exhaustive endgame computation feasible while retaining nontrivial branching and transpositions.

\subsubsection{Terminal Utility for Othello}
\label{section:rule-othello}

The full rules are given in Appendix~\hyperref[app:rules-othello]{B.1}.

\paragraph{Terminal utility used in our experiments.}
When the game ends with empty squares remaining, we assign all remaining empty squares to the winner for scoring purposes.
Equivalently, the winner's score margin is increased by the number of empty squares (and the loser's margin decreased by the same amount), while a draw remains a draw.
Formally, for a terminal position \(p\), let \(\Delta(p)\) denote this final score margin (a signed integer) from Black's perspective.
We define the terminal utility \(\texttt{value}(p)\) under the negamax convention as the margin from the perspective of the side to move at \(p\): \(\texttt{value}(p)=\Delta(p)\) if the side to move is Black, \(\texttt{value}(p)=-\Delta(p)\) if the side to move is White, and \(\texttt{value}(p)=0\) in case of a draw.

\subsubsection{Position Counts for \(6\times 6\) Othello}
\label{sec:othello-position-counts}

To quantify structural coverage, we measured the number of \emph{distinct positions} encountered under three regimes:
(i) weak solving via alpha--beta,
(ii) semi-strong solving via reopening alpha--beta, and
(iii) (partial) strong enumeration via exhaustive breadth-first search (BFS) over rule-reachable positions.
For alpha--beta based regimes, a position is counted when it is \emph{expanded} at least once (i.e., when legal moves are generated and searched from that position under our counting conventions).
For BFS, a position is counted when it is \emph{enumerated} as rule-reachable from \(p_0\) under arbitrary play.
Selected disc counts and the overall totals are reported in Table~\ref{tab:retro_result_66othello}; Table~\ref{tab:retro_result_66othello_full} gives the complete results.

\paragraph{Canonicalization and counting conventions.}
All counts are taken over canonical representatives of equivalence classes of positions.
First, we identify positions up to the eight dihedral symmetries of the square board (rotations and reflections)~\cite{schiffel2010symmetry}.
Second, we identify positions up to simultaneous color inversion and side-to-move inversion (swap Black/White discs and swap the player to move); under the negamax convention and our terminal utility, this preserves values from the perspective of the side to move.
Third, we count only positions at which the side to move has at least one non-pass legal move; positions whose unique legal move is the forced pass are excluded from all counts.
Equivalently, we contract forced-pass transitions for reporting, and we apply the same convention uniformly across regimes.

\paragraph{Determinism, PV fixing, and what is (and is not) counted.}
Both alpha--beta based solvers use deterministic control decisions.
In particular, at PV nodes we fix the principal move to follow the well-known perfect-play line for \(6\times 6\) Othello (Feinstein's perfect play)~\cite{Feinstein1993Amenor,takeshita2016othello66}.
This isolates the overhead due to certification obligations from variability due to PV discovery, and ensures that the principal branch is identical across the weak and semi-strong runs.

The weak counts cover the main alpha--beta/PVS procedure; the semi-strong counts cover the obligation-enforcing certification phase, including retained certified positions and auxiliary cut/all positions.
Auxiliary move-selection searches are excluded from both regimes. Appendix~\hyperref[app:othello-selectors]{C.1} gives the selection procedures and move-ordering heuristic.
Each regime contributing to Table~\ref{tab:retro_result_66othello} was executed once with the stated PV convention and implementation; the reported counts are from that run.

\paragraph{Stored Positions and a Fixed Deployment Policy.}
Let \(S\) denote the retained position representatives carrying a certified kind \(P\), \(A'\), or \(P'\), under the counting conventions above.
The solution-artifact column reports \(|S|=97{,}252{,}321\) records after aggregation; these may include additional certified records alongside a fixed deployment policy's region.
Appendix~\hyperref[app:othello-policy]{C.2} describes how the retained successor obligations support that policy.

\begin{table}[htbp]
\caption{Selected disc counts for $6\times6$ Othello. Totals include all disc counts, including rows omitted here; the complete results are in Table~\ref{tab:retro_result_66othello_full} (Appendix~\hyperref[app:complete-counts]{D}). Counts use the conventions of Section~\ref{sec:othello-position-counts}; auxiliary move-selection searches are excluded. The solution artifact is the retained certified set $S$. Daggers mark unfinished strong enumeration.}
\label{tab:retro_result_66othello}
\centering
\small
\begin{tabular}{r|r|r|r|r|r}
\hline
category & weak & \multicolumn{3}{c|}{semi-strong} & strong\\\hline
regime & alpha--beta & \multicolumn{3}{c|}{reopening alpha--beta} & exhaustive BFS\\\hline
component & (total) & solution artifact & proof certificate & total & (total)\\\hline\hline
\multicolumn{1}{r|}{discs} & \multicolumn{5}{c}{number of positions} \\\hline
4 & 1 & 1 & 0 & 1 & 1 \\
8 & 12 & 12 & 20 & 32 & 60 \\
12 & 287 & 343 & 2,121 & 2,464 & 51,964 \\
16 & 6,864 & 10,015 & 132,731 & 142,746 & 51,072,917 \\
20 & 177,787 & 225,426 & 6,045,608 & 6,271,034 & 19,791,417,568 \\
25 & 8,819,433 & 3,896,228 & 390,330,164 & 394,226,392 & 2,417,685,025,700 \\
26 & 15,263,415 & 6,035,411 & 832,468,650 & 838,504,061 & $\dagger$ \\
32 & 240,828,705 & 10,736,694 & 7,894,710,441 & 7,905,447,135 & $\dagger$ \\
35 & 90,900,623 & 1,255,057 & 1,003,741,866 & 1,004,996,923 & $\dagger$ \\
\hline
total & 1,339,211,721 & 97,252,321 & 42,889,981,900 & 42,987,234,221 & $\gg$ 4,473,922,545,824 \\\hline
\end{tabular}
\end{table}

\paragraph{Overall overhead of semi-strong certification.}
As shown in Table~\ref{tab:retro_result_66othello}, semi-strong solving expands \(42{,}987{,}234{,}221\) distinct positions in the primary certification phase, whereas weak solving expands \(1{,}339{,}211{,}721\) distinct positions in the primary alpha--beta phase.
Thus, semi-strong certification expands approximately \(32.1\times\) as many distinct positions as weak solving under our counting definition.
This empirical ratio does not directly test the idealized tree bound. There are at most 32 disc placements after the initial board, but the formal move depth can additionally include forced passes; the experiment also has variable branching, transpositions, and excluded auxiliary probes.

\paragraph{Deployability vs.\ verifiability.}
The semi-strong total decomposes into a small deployed artifact and a much larger verification component.
The \emph{solution artifact} retains \(|S|=97{,}252{,}321\) expanded positions (about \(0.23\%\) of the semi-strong total), supporting exact-value queries and a fixed policy as described in Appendix~\hyperref[app:othello-policy]{C.2}.
The remaining \(42{,}889{,}981{,}900\) expanded positions are retained only for the \emph{proof certificate} as auxiliary cut/all evidence.
This separation emphasizes that deployability can be achieved with a compact artifact even when full third-party verifiability requires substantially more auxiliary information.

\paragraph{Rule-reachable positions and the gap to strong solving.}
We also attempted to enumerate rule-reachable positions under arbitrary play via exhaustive BFS.
Due to storage limitations, this enumeration could only be completed through positions with 25 discs on the board.
Even this partial BFS already enumerates at least \(4{,}473{,}922{,}545{,}824\) distinct rule-reachable positions (Table~\ref{tab:retro_result_66othello}, strong total), which is over \(100\times\) larger than the total number of positions expanded by semi-strong solving for the full game.
This illustrates why strong solving by naive full enumeration is substantially more demanding.

\paragraph{Why weak alpha--beta expands more positions than the deployed artifact stores.}
The deployed solution artifact retains certified positions sufficient for value and optimal-move queries in the supported region. Its retained size need not be a minimum representation.
By contrast, even when the PV is fixed, a weak alpha--beta computation expands additional auxiliary positions (in particular cut/all positions) to establish pruning bounds and to justify the root value.
Crucially, the effectiveness of alpha--beta in limiting this auxiliary expansion depends on move ordering, i.e., whether high-scoring moves are tried early enough to trigger cutoffs.
Game-solving search can encounter positions unlike those seen in competitive play. Distribution shift has been studied for learned search heuristics in another game-solving setting~\cite{online-fine-tuning-NEURIPS2023}; its contribution to the present Othello counts was not measured.
Moreover, at cut nodes the first explored move is required only to \emph{fail high} (reach or exceed \(\beta\)), and need not be the canonical best move that would be selected under perfect play.
As a result, the set of positions expanded by weak alpha--beta can be substantially larger than the set of positions that must be stored for deployment, even though the solution artifact alone already suffices to recover a weak solution (Proposition~\ref{prop:artifact-implies-weak}).

\paragraph{Interpretation via Othello's pruning and mobility characteristics.}
Viewed through the lens of weak solving, the solution artifact supports play in each derived game obtained by fixing one player's canonical optimal policy and allowing arbitrary choices by the other player.
One possible explanation of the artifact/proof gap is that a move may suffice for a cutoff without maximizing the score margin, while the canonical continuation may lead to a different mobility profile. This explanation is a conjecture; the experiment does not isolate either effect.

\paragraph{Artifact release and verifiability.}
GPW-14 reports constructing a strongly solved game tree for \(6\times 6\) Othello on a volunteer/desktop-grid computing system, by splitting the search into subtrees~\cite{GPW2014-6x6othello}.
To the best of our knowledge, the corresponding IPSJ Digital Library record provides the paper PDF only and does not include additional downloadable solution artifacts (e.g., a transposition-reduced state-value/move database or search traces)~\cite{GPW2014-6x6othello}.
Accordingly, quantities such as the size of the deduplicated rule-reachable state space are not directly verifiable from public artifacts.
In contrast, our semi-strong solution artifact is compact enough to be deposited on Zenodo; see Appendix~E (Online Resources).
For reference, the full public Zenodo bundle for the \(6\times 6\) Othello release occupies 138.4\,GB.
This figure documents the size of the released objects for this experiment; it is not intended as a cross-game benchmark.
This public release facilitates independent verification and downstream comparisons, supporting the view that semi-strong solving offers a practical compromise between deployment efficiency and verifiability.
Concretely, the released \(6\times 6\) Othello tablebase stores exact values on \(S\) together with node-kind information that supports the fixed deployment policy described in Appendix~\hyperref[app:othello-policy]{C.2}. It supplies exact values on that policy's certified region \(R\) and canonical move extraction at its optimal-agent decision points.
The accompanying Zenodo release fixes this tablebase in a versioned, citable form for independent verification and downstream comparison.

\subsection{Semi-Strongly Solving Connect Four}

We additionally evaluate our framework on standard Connect Four (7 columns \(\times\) 6 rows).

\subsubsection{Experimental Setup}
Unlike the \(6\times 6\) Othello experiment, Connect Four is already strongly solved in the literature~\cite{boeck2025connect4strong}, and the public strong-solution dataset/probing tables used by our reproduction code are available on Zenodo~\cite{boeck2025connect4zenodo}. By utilizing them, we carry out a controlled ``oracle'' experiment:
we assume access to the exact game-theoretic value of any queried position and use that information to perform value-optimal move ordering.
The purpose of this experiment is not to re-solve Connect Four, but to measure the \emph{structural sizes} of (i) the deployed semi-strong solution artifact and (ii) the associated proof certificate under ideal move ordering, and to compare them to the size of the strong (rule-reachable) state space.
Accordingly, our public reproduction package uses the existing public strong-solution source as an oracle and does not redundantly re-publish a copy of the upstream strong tablebase.
What we release for this experiment is the code and machine-readable aggregate results needed to reproduce the measurements reported here.

\paragraph{Terminal utility and value granularity.}
We use the 3-valued WDL utility (\(\textsc{Win}/\textsc{Draw}/\textsc{Loss}\)) as the terminal utility.
In particular, we do \emph{not} refine values by depth-to-win or depth-to-loss (we do not prefer ``faster wins'' among winning lines).
This choice reflects the value information available from the strong-solution source used for the oracle and intentionally contrasts with the integer score utility used in the Othello experiment.

\paragraph{Tie-breaking in move ordering.}
When multiple legal moves attain the same maximal backed-up value, we break ties deterministically by preferring moves closer to the center column, and then preferring the left side.
This rule instantiates the global tie-breaking rule \(\mathrm{TB}\) (Section~\ref{sec:game-model}) for Connect Four and thereby fixes the canonical move \(m^{\star}(\cdot)\).

\paragraph{Symmetry convention for counting.}
The Connect Four board admits a left--right reflection symmetry.
However, unlike the Othello experiment, we do \emph{not} quotient positions by this symmetry; symmetric positions are counted separately.
The reason is that our strong (rule-reachable) position counts are taken from published values that follow the same convention, and we match that convention for comparability.

\paragraph{Disc count and terminal positions.}
Connect Four has no passes, hence the side to move is determined by the disc count (parity).
Unlike the Othello experiment, we \emph{include terminal positions} in the counts; these are visited or recorded but do not generate successors; in particular, the disc-count \(42\) row corresponds to full-board terminal positions, including draws and wins completed on the 42nd move (and other terminal positions at earlier disc counts are also included).

\subsubsection{Terminal Utility for Connect Four}
\label{section:rule-connect4}

The full rules are given in Appendix~\hyperref[app:rules-connect4]{B.2}.

\paragraph{Terminal utility (WDL)}
For a terminal position \(p\), we define \(\texttt{value}(p)\in\{-1,0,+1\}\) under the negamax convention as follows:
\(\texttt{value}(p)=+1\) if the side to move at \(p\) has already won (this case does not arise under normal play),
\(\texttt{value}(p)=-1\) if the side to move has already lost (i.e., the previous move created four in a row for the opponent),
and \(\texttt{value}(p)=0\) if the game is a draw.
The game-theoretic value \(V(\cdot)\) is induced by this terminal utility as in Section~\ref{sec:game-model}.

\subsubsection{Position Counts and Artifact Sizes}
\label{sec:connect4-position-counts}

To compare weak, semi-strong, and strong notions quantitatively, Table~\ref{tab:retro_result_connect4} reports selected disc counts and the overall totals; Table~\ref{tab:retro_result_connect4_full} gives the complete results.
For weak solving, the reported counts correspond to the number of distinct positions visited by a conventional alpha--beta/PVS procedure.
For semi-strong solving, we report both the deployed solution-artifact component (positions retained for deployment) and the additional proof-certificate component (auxiliary cut/all positions required for verification), together with their total.
For strong solving, we report the total number of rule-reachable positions under arbitrary play (from published counts), using the same convention of counting left--right symmetric positions separately.

\paragraph{Oracle Move Ordering.}
Both procedures order moves at every expanded position, including cut nodes, by decreasing exact WDL value, using the fixed tie-breaking rule above.
Each regime was executed once. The complete oracle procedure and its scope are described in Appendix~\hyperref[app:connect4-oracle]{C.3}.

\begin{table}[htbp]
\caption{Selected disc counts for Connect Four. Totals include all disc counts, including rows omitted here; the complete results are in Table~\ref{tab:retro_result_connect4_full} (Appendix~\hyperref[app:complete-counts]{D}). Counts use oracle WDL ordering and fixed tie-breaking, include terminal positions, and count reflected positions separately (Section~\ref{sec:connect4-position-counts}).}
\label{tab:retro_result_connect4}
\centering
\small
\begin{tabular}{r|r|r|r|r|r}
\hline
category & weak & \multicolumn{3}{c|}{semi-strong} & strong\\\hline
regime & alpha--beta & \multicolumn{3}{c|}{reopening alpha--beta} & rule-reachable\\\hline
component & (total) & solution artifact & proof certificate & total & (total)\\\hline\hline
\multicolumn{1}{r|}{discs} & \multicolumn{5}{c}{number of positions} \\\hline
0 & 1 & 1 & 0 & 1 & 1 \\
1 & 1 & 7 & 0 & 7 & 7 \\
6 & 260 & 477 & 534 & 1,011 & 16,422 \\
12 & 12,442 & 25,741 & 34,093 & 59,834 & 12,236,101 \\
18 & 313,016 & 667,698 & 721,836 & 1,389,534 & 1,763,883,894 \\
24 & 2,857,888 & 6,816,505 & 5,246,286 & 12,062,791 & 60,830,813,459 \\
30 & 7,927,180 & 22,627,524 & 12,442,361 & 35,069,885 & 410,378,505,447 \\
36 & 5,843,979 & 22,724,558 & 9,178,727 & 31,903,285 & 266,313,901,222 \\
42 & 0 & 1,683,726 & 441,663 & 2,125,389 & 1,459,332,899 \\
\hline
total & 96,450,672 & 342,134,570 & 157,309,404 & 499,443,974 & 4,531,985,219,092 \\\hline
\end{tabular}
\end{table}

\paragraph{Semi-strong vs.\ strong.}
Table~\ref{tab:retro_result_connect4} shows that semi-strong certification constructs a proof (artifact plus certificate) over \(499{,}443{,}974\) distinct positions, whereas the strong (rule-reachable) state space contains \(4{,}531{,}985{,}219{,}092\) positions.
Under this oracle WDL ordering and fixed tie-breaking rule, the semi-strong certification structure therefore contains approximately \(9{,}074\) times fewer distinct positions than the rule-reachable baseline.

\paragraph{Semi-strong vs.\ weak.}
Semi-strong certification visits \(499{,}443{,}974\) positions in total, while weak solving visits \(96{,}450{,}672\), a factor of approximately \(5.18\times\).
The weak procedure constructs a proof of the initial value and an associated strategy; semi-strong certification additionally requires exact value queries and canonical move extraction over its certified region. Neither count is simply the number of root values established.
The disc-count \(1\) row provides a simple illustration: weak solving visits only the single PV successor from the initial position, while semi-strong certification must cover all seven legal first moves because the initial position is a free-agent choice point in one of the two orientations defining \(R\).

\paragraph{Artifact vs.\ certificate.}
The semi-strong total decomposes into \(342{,}134{,}570\) positions retained in the deployed solution artifact and \(157{,}309{,}404\) additional positions retained only for the proof certificate.
Thus, the artifact accounts for approximately \(68.5\%\) of the semi-strong total in this experiment.
This contrasts with Othello, where auxiliary proof positions dominate. The two experiments differ in game structure, utility granularity, tie-breaking, and move ordering, so these measurements do not isolate the contribution of any one factor.

\paragraph{Remarks on tie-breaking and symmetry.}
Because WDL values are coarse, ties among value-maximizing moves are common. Different tie-breaking rules can change \(m^{\star}(\cdot)\), and hence \(R\) and the stored artifact; the present experiment uses only one rule and does not measure this sensitivity.
Our deterministic center-first, then left-first rule fixes this dependence.
We also note that quotienting by left--right reflection symmetry would reduce counts, but not by an exact factor of two due to self-symmetric positions; we avoid symmetry quotienting here to remain comparable to the published strong counts.

\section{Discussion}
\label{sec:discussion}

\paragraph{When the Additional Coverage Is Useful.}
Semi-strong solving keeps the original minimax value function and changes the certification target. A weak solution already guarantees the initial result against arbitrary replies; semi-strong solving additionally requires exact local values and canonical optimal responses throughout the specified region. This matters when an engine should exploit newly available wins or score gains, or when a user wants exact analysis of a game played against that engine. In human--AI play, it suffices for the AI to follow its canonical policy; the human may make mistakes on any number of turns.

The cost is additional construction and storage, and the scope remains limited. Weak solving is sufficient when only the initial perfect-play result and a strategy securing it are wanted. Strong solving remains the appropriate guarantee for arbitrary starting positions or play in which both players depart from the supported policy. Semi-strong solving is useful when exact analysis of its narrower deployment region is worth the additional resources; it does not promise coverage of most rule-reachable positions.

\paragraph{Construction Cost and Stored Information.}
The Othello artifact supports deployment with far fewer retained positions than the auxiliary proof component, whereas the Connect Four artifact constitutes a larger fraction of its certification structure. A verifier that recomputes auxiliary reasoning also trades stored evidence for computation.

End-to-end wall-clock and aggregate CPU-time records for constructing the reported artifacts are unavailable, so we cannot report those totals. A reproducible cost comparison would separately report move selection/probes, certification, deduplication/export, peak memory, and any upstream-oracle cost, with wall-clock time distinguished from summed CPU time.

\paragraph{Tie-Breaking and Reproducibility.}
The terminal utility determines \(V\); a fixed tie-breaking rule selects a policy and therefore determines \(R\). Different engines can choose different tied optimal moves, so their artifacts need not contain the same positions even when all values agree. Reproducing a retained artifact requires matching search decisions, cache context, canonicalization, and representation conventions. For Othello, Section~\ref{sec:othello-position-counts} distinguishes the retained set \(S\) from a fixed policy's region; Appendix~\hyperref[app:othello-policy]{C.2} explains how certified successors support a consistent deployment policy. A different tied move is supported only if its required continuation is covered. Connect Four instead uses the explicit position-based tie rule specified in Section~4.2.
The experiments do not quantify how much artifact size changes with \(\mathrm{TB}\). A structure-aware rule could choose among equally optimal moves using an estimate of continuation size or future free-agent mobility. Such a rule must first preserve the minimax value. Heuristic size estimates may avoid fully expanding every tied alternative, but they do not guarantee a minimum-size artifact. Exact minimization is more difficult because transpositions make shared continuation costs non-additive. Comparing fixed positional rules with such heuristics is a natural future experiment, rather than a result established here.

\paragraph{Search Strategies Beyond Alpha--Beta.}
The definition of semi-strong solving and its recursive coverage requirements do not depend on depth-first traversal. A best-first or proof-number-style procedure could instead schedule unresolved value and coverage obligations, retaining all free-agent successors and the certified canonical continuation at optimal-agent turns. It would also have to distinguish exact values from one-sided bounds and track unresolved obligations through transpositions. We do not implement such a variant here, and the node-count bound proved for reopening alpha--beta does not automatically transfer to it.

\paragraph{Other Extensions.}
The present model assumes finite, alternating-turn, two-player, zero-sum, perfect-information games without chance. Stochastic or imperfect-information settings would require new definitions and proofs. Within the present model, useful extensions include additional benchmark games and more compact certificates with incremental or out-of-core verification.

\section{Conclusions}
\label{sec:conclusions}

Semi-strong solving certifies exact values and canonical optimal decisions on the region reached when one fixed player follows its canonical optimal policy and the opponent may play arbitrarily, for either player orientation. This strengthens the local information required by weak solving while retaining a potentially smaller target than all rule-reachable positions. Reopening alpha--beta implements these requirements with explicit node obligations and has an \(O(d\,b^{d/2})\) total node bound in the fixed-branching, ideal-ordering tree model.

The Othello experiment constructs a deployable solution artifact while exhaustive enumeration reaches a multi-trillion position count before exhausting storage. The Connect Four experiment, using oracle WDL ordering and a fixed tie-breaking rule, produces a certification structure with approximately \(9{,}074\) times fewer positions than the published rule-reachable baseline. These results demonstrate structural coverage and storage trade-offs. Quantitative sensitivity to tie-breaking remains to be established.

\section*{Competing Interests}
Competing interests: The author is employed by Preferred Networks Inc. The author declares that this affiliation did not influence the research outcomes.

\printbibliography

\appendix
\section{Detailed Proofs for Selected Section~3 Results}
\label{app:selected-proofs}

This appendix provides detailed proofs for Propositions~\ref{prop:derived-game-view},
\ref{prop:artifact-implies-weak}, \ref{prop:node-kind-correspondence}, and~\ref{prop:tt-reuse},
and for Lemmas~\ref{lem:AprimePprime} and~\ref{lem:CA}.

\subsection{Detailed Proof of Proposition~\ref{prop:derived-game-view}}
\label{app:proof-derived-game-view}
Let $\mathrm{Reach}_P$ denote the set of positions reachable from $p_0$ in the derived game $G_P$ of Definition~\ref{def:derived-game}.
By definition of reachability, $\mathrm{Reach}_P$ is the smallest set $S$ of positions such that
\begin{enumerate}
\item $p_0\in S$;
\item if $p\in S$ and it is the free agent's turn at $p$, then $p\cdot m\in S$ for every $m\in M(p)$; and
\item if $p\in S$ is non-terminal and it is the optimal agent's turn at $p$, then $p\cdot m^{\star}(p)\in S$.
\end{enumerate}
These are exactly the three clauses used in Section~\ref{sec:certified-region} to define $R_P$.
Hence
\[
\mathrm{Reach}_P = R_P.
\]

Now let $V_{G_P}(p)$ denote the game-theoretic value of $p$ in the derived game $G_P$.
We claim that
\[
V_{G_P}(p)=V(p) \qquad\text{for every } p\in R_P.
\]
We prove this by induction on the remaining play length from $p$ in $G_P$.
If $p$ is terminal, then $G_P$ has the same terminal utility as the original game, so
\[
V_{G_P}(p)=\texttt{value}(p)=V(p).
\]
Assume now that $p$ is non-terminal and that the claim holds for all successors of $p$ in $G_P$.
There are two cases.

\emph{Case 1: the free agent is to move at $p$.}
In this case, the legal moves in $G_P$ are exactly the original legal moves $M(p)$.
Therefore,
\[
V_{G_P}(p)=\max_{m\in M(p)} -V_{G_P}(p\cdot m).
\]
By the induction hypothesis, $V_{G_P}(p\cdot m)=V(p\cdot m)$ for every such successor, so
\[
V_{G_P}(p)=\max_{m\in M(p)} -V(p\cdot m)=V(p)
\]
by the defining negamax recursion of Section~\ref{sec:game-model}.

\emph{Case 2: the optimal agent is to move at $p$.}
In $G_P$, the legal-move set is the singleton $\{m^{\star}(p)\}$.
Thus
\[
V_{G_P}(p)=-V_{G_P}(p\cdot m^{\star}(p)).
\]
Applying the induction hypothesis to the unique successor gives
\[
V_{G_P}(p)=-V(p\cdot m^{\star}(p)).
\]
By definition of the canonical optimal move,
\[
-V(p\cdot m^{\star}(p))=
\max_{m\in M(p)} -V(p\cdot m)=V(p).
\]
Hence $V_{G_P}(p)=V(p)$ also in this case.
This proves the claim.

Since $\mathrm{Reach}_P=R_P$ and the value functions agree on that set, determining the value component on $R_P$ is exactly the same task as determining the game-theoretic value of every position reachable from $p_0$ in $G_P$.
Therefore statements~(1) and~(2) of Proposition~\ref{prop:derived-game-view} are equivalent.
\hfill$\square$

\subsection{Detailed Proof of Proposition~\ref{prop:artifact-implies-weak}}
\label{app:proof-artifact-implies-weak}
If $p_0$ is terminal, then by definition of a solution artifact the artifact $\mathcal A$ returns
\[
V(p_0)=\texttt{value}(p_0),
\]
and the required strategy is vacuous.
So only the non-terminal case requires argument.

Assume that $p_0$ is non-terminal.
Because $p_0\in R$ by definition of the certified region, $\mathcal A$ returns the exact value $V(p_0)$.
It remains to construct a strategy for the first player that guarantees at least this value from the start.

Consider the orientation in which the \emph{first} player is the optimal agent, and write $G_{\text{first}}$ for the corresponding derived game of Definition~\ref{def:derived-game}.
For every non-terminal position $p\in R_{\text{first}}$ at which the first player is to move, define
\[
\sigma(p):=m^{\star}(p),
\]
where $m^{\star}(p)$ is obtained from $\mathcal A$.
This is well-defined by the assumption that the artifact returns the canonical move at optimal-agent turns in the relevant orientation.

We first show that any play starting from $p_0$ in which the first player follows $\sigma$ remains inside $R_{\text{first}}$.
This is immediate by induction on the play length.
At the beginning, $p_0\in R_{\text{first}}$ by clause~(1) of the definition.
If a current position $p\in R_{\text{first}}$ is the second player's turn, then every legal move $m\in M(p)$ satisfies
\[
p\cdot m\in R_{\text{first}}
\]
by clause~(2).
If instead it is the first player's turn, then $\sigma(p)=m^{\star}(p)$, so the successor
\[
p\cdot \sigma(p)=p\cdot m^{\star}(p)
\]
lies in $R_{\text{first}}$ by clause~(3).

Now compare this play rule to the derived game $G_{\text{first}}$.
By Proposition~\ref{prop:derived-game-view}, the reachable positions of $G_{\text{first}}$ are exactly $R_{\text{first}}$, and the game-theoretic value in $G_{\text{first}}$ agrees with the original value $V(\cdot)$ on that set.
Moreover, in $G_{\text{first}}$, whenever the first player is to move at $p\in R_{\text{first}}$, the legal-move set is exactly
\[
\{m^{\star}(p)\}=\{\sigma(p)\}.
\]
Thus the first player's legal strategy in $G_{\text{first}}$ is precisely the strategy $\sigma$.

Because the value of $G_{\text{first}}$ at the initial position equals $V(p_0)$, the minimax meaning of $V_{G_{\text{first}}}(p_0)$ implies that the strategy $\sigma$ guarantees a first-player payoff of at least $V(p_0)$ against arbitrary second-player replies in $G_{\text{first}}$, with equality against an optimal opponent.
Finally, every play of the original game in which the first player follows $\sigma$ is also a play of $G_{\text{first}}$, and conversely every play of $G_{\text{first}}$ is a play of the original game consistent with $\sigma$.
The terminal utilities are the same in both games.
Therefore the same strategy $\sigma$ guarantees a first-player payoff of at least $V(p_0)$ against arbitrary second-player replies in the original game as well.

Hence $\mathcal A$ determines both the exact initial value and a strategy from the start that guarantees at least that value.
This is exactly a weak solution in the usual sense.
\hfill$\square$

\subsection{Detailed Proof of Lemma~\ref{lem:AprimePprime}}
\label{app:proof-AprimePprime}
Write
\[
a_d := N(A',d),
\qquad
p_d := N(P',d).
\]
From Section~\ref{section:generation-rule},
\[
a_1=0,
\qquad
p_1=0,
\qquad
N(P,d)=1\quad(d\ge 1),
\]
and for every $d>1$,
\[
a_d=(b-1)N(P,d-1)+b\,p_{d-1},
\qquad
p_d=a_{d-1}.
\]
Substituting $N(P,d-1)=1$ and $p_{d-1}=a_{d-2}$ gives, for every $d>2$,
\[
a_d=(b-1)+b\,a_{d-2}.
\]
Also,
\[
a_2=(b-1)N(P,1)+b\,p_1=b-1.
\]
Thus the $A'$ counts satisfy the second-order recurrence
\[
a_1=0,
\qquad
a_2=b-1,
\qquad
a_d=(b-1)+b\,a_{d-2}\quad(d>2).
\]

We now solve this recurrence separately on even and odd levels.
For $k\ge 1$, define $e_k:=a_{2k}$.
Then $e_1=a_2=b-1$, and for $k\ge 2$,
\[
e_k=(b-1)+b\,e_{k-1}.
\]
Unrolling gives
\[
e_k
=(b-1)\sum_{i=0}^{k-1} b^i
= b^k-1.
\]
Therefore,
\[
a_{2k}=b^k-1 \qquad (k\ge 1).
\]

For $k\ge 0$, define $o_k:=a_{2k+1}$.
Then $o_0=a_1=0$, and for $k\ge 1$,
\[
o_k=(b-1)+b\,o_{k-1}.
\]
Again unrolling gives
\[
o_k
=(b-1)\sum_{i=0}^{k-1} b^i
= b^k-1.
\]
Hence,
\[
a_{2k+1}=b^k-1 \qquad (k\ge 0).
\]

Combining the even and odd cases,
\[
N(A',d)=a_d=b^{\left\lceil\frac{d-1}{2}\right\rceil}-1
\qquad (d\ge 1).
\]
Finally, for $d>1$ we have $p_d=a_{d-1}$, so
\[
N(P',d)=p_d=b^{\left\lceil\frac{d-2}{2}\right\rceil}-1.
\]
This same formula also gives $N(P',1)=0$, because $\lceil(1-2)/2\rceil=0$.
Thus the stated closed forms hold for all $d\ge 1$.
\hfill$\square$

\subsection{Detailed Proof of Lemma~\ref{lem:CA}}
\label{app:proof-CA}
Write
\[
c_d := N(C,d),
\qquad
u_d := N(A,d),
\qquad
q_d := N(A',d).
\]
From the recurrences in Section~\ref{section:generation-rule},
\[
c_1=0,
\qquad
u_1=0,
\qquad
u_d=c_{d-1} \quad(d>1),
\]
and for $d>1$,
\[
c_d=(b-1)q_{d-1}+b\,u_{d-1}.
\]
Using $u_{d-1}=c_{d-2}$ for $d>2$, we obtain
\[
c_d=(b-1)q_{d-1}+b\,c_{d-2}
\qquad(d>2).
\]
By Lemma~\ref{lem:AprimePprime},
\[
q_{d-1}=b^{\left\lceil\frac{d-2}{2}\right\rceil}-1.
\]
Therefore, for every $d>2$,
\[
c_d=(b-1)\left(b^{\left\lceil\frac{d-2}{2}\right\rceil}-1\right)+b\,c_{d-2}.
\]
We also need the base values
\[
c_2=(b-1)q_1+b\,u_1=0.
\]

We again split by parity.
For even levels, define $e_k:=c_{2k}$ for $k\ge 1$.
Then $e_1=c_2=0$, and for $k\ge 2$,
\[
e_k=(b-1)(b^{k-1}-1)+b\,e_{k-1}.
\]
Unrolling this first-order recurrence gives
\[
e_k
=\sum_{i=2}^{k} b^{k-i}(b-1)(b^{i-1}-1).
\]
We simplify the sum term by term:
\begin{align*}
e_k
&=(b-1)\sum_{i=2}^{k} b^{k-i}b^{i-1}
  -(b-1)\sum_{i=2}^{k} b^{k-i}\\
&=(b-1)\sum_{i=2}^{k} b^{k-1}
  -(b-1)\sum_{j=0}^{k-2} b^j\\
&=(k-1)(b-1)b^{k-1}-(b^{k-1}-1)\\
&=(k-1)b^k-kb^{k-1}+1.
\end{align*}
Hence
\[
c_{2k}=(k-1)b^k-kb^{k-1}+1.
\]

For odd levels, define $o_k:=c_{2k+1}$ for $k\ge 0$.
Then $o_0=c_1=0$, and for $k\ge 1$,
\[
o_k=(b-1)(b^k-1)+b\,o_{k-1}.
\]
Unrolling gives
\[
o_k=\sum_{i=1}^{k} b^{k-i}(b-1)(b^{i}-1).
\]
Again simplify:
\begin{align*}
o_k
&=(b-1)\sum_{i=1}^{k} b^{k-i}b^i
  -(b-1)\sum_{i=1}^{k} b^{k-i}\\
&=(b-1)\sum_{i=1}^{k} b^k
  -(b^k-1)\\
&=k(b-1)b^k-(b^k-1)\\
&=kb^{k+1}-(k+1)b^k+1.
\end{align*}
Therefore
\[
c_{2k+1}=kb^{k+1}-(k+1)b^k+1.
\]

The even and odd formulas are exactly the compact ceiling-expression
\[
N(C,d)=
\left\lceil \frac{d-2}{2} \right\rceil b^{\left\lceil \frac{d}{2} \right\rceil}
-\left\lceil \frac{d}{2} \right\rceil b^{\left\lceil \frac{d-2}{2} \right\rceil}+1,
\qquad d\ge 1.
\]
Indeed, for $d=2k$ it becomes $(k-1)b^k-kb^{k-1}+1$, and for $d=2k+1$ it becomes $kb^{k+1}-(k+1)b^k+1$.

Finally, for $d>1$ we have $N(A,d)=u_d=c_{d-1}$, so substituting $d-1$ into the formula for $c_{d-1}$ yields
\[
N(A,d)=
\left\lceil \frac{d-3}{2} \right\rceil b^{\left\lceil \frac{d-1}{2} \right\rceil}
-\left\lceil \frac{d-1}{2} \right\rceil b^{\left\lceil \frac{d-3}{2} \right\rceil}+1
\qquad (d>1).
\]
For $d=1$, both sides equal $0$ by direct inspection.
Hence the stated formula for $N(A,d)$ holds for all $d\ge 1$.
\hfill$\square$

\subsection{Detailed Proof of Proposition~\ref{prop:node-kind-correspondence}}
\label{app:proof-node-kind-correspondence}
Because there are no transpositions and no repetitions, every recursive call of Algorithm~\ref{tab:alg2} reaches a distinct successor position, so the visited structure is an ordinary rooted tree.
It therefore suffices to characterize, for each node kind, which recursive child calls are made.

Under the assumptions of optimal move ordering and distinct backed-up values, the first move in the deterministic order is the unique eventual principal move at every nonterminal node.
Consequently, whenever the loop in Algorithm~\ref{tab:alg2} explores more than one child, the first child is the eventual PV child and every later child is non-PV.
At certified nodes, the first child sets $\alpha$ to the exact best-child value. At auxiliary nodes, the ideal cut/all invariant ensures that no later child raises $\alpha$. Hence the PVS re-search branch at lines~17--19 is never taken. A fail-low all-node need not set $\alpha$ equal to its exact value.

We now inspect the control flow by node kind.

\emph{$P$-nodes.}
At a $P$-node, Algorithm~\ref{tab:alg2} resets the window to $[-\infty,+\infty]$ (lines~3--5), and beta-cutoff is disabled because line~22 applies only when $k\notin\{\texttt{"P"},\texttt{"P'"}\}$.
Hence all legal moves are searched.
The first child is the unique principal child, so Algorithm~\ref{tab:alg1} assigns it kind $P$.
Every later child is non-principal, so Algorithm~\ref{tab:alg1} assigns it kind $A'$.
Thus a visited edge out of a $P$-node follows exactly the rule stated in Section~\ref{section:generation-rule}.

\emph{$A'$-nodes.}
At an $A'$-node, the window is again reopened to $[-\infty,+\infty]$ by lines~3--5.
The first child is the unique principal child and is therefore searched with kind $P'$ by Algorithm~\ref{tab:alg1}.
Every later child is first searched under the non-PV kind $C$ via line~16.
Because the first child is already the unique best child, each later child has score at most the current $\alpha$, so no later child is promoted to PV and the re-search branch at lines~17--19 never fires.
Hence the principal child of an $A'$-node is a $P'$-node and all non-principal visited children are $C$-nodes.

\emph{$P'$-nodes.}
At a $P'$-node, the window is reopened to $[-\infty,+\infty]$ and beta-cutoff is disabled for the same reason as at a $P$-node.
Therefore all legal moves are searched.
Algorithm~\ref{tab:alg1} returns kind $A'$ for every child of a $P'$-node, regardless of whether it is principal.
Thus every visited child of a $P'$-node is an $A'$-node.

\emph{$C$- and $A$-nodes.}
After the cases above, the remaining recursive calls occur only inside the non-principal subtrees entered through $C$-nodes.
For these kinds, Algorithm~\ref{tab:alg2} behaves exactly as ordinary alpha--beta with the kind transitions given by Algorithm~\ref{tab:alg1}: a $C$-node always passes kind $A$ to its visited child, and an $A$-node always passes kind $C$ to each child it searches.
Under optimal move ordering and distinct values, standard alpha--beta yields Knuth's minimal cut/all structure on such non-PV subtrees: each cut node visits exactly one child (fail-high), and each all node visits all children (fail-low)~\cite{knuth1975analysis-alphabeta}.
Therefore each visited $C$-node has exactly one visited child, of kind $A$, and every visited child of an $A$-node is of kind $C$.

Collecting the four cases shows that every visited edge follows the kind transition stated in Section~\ref{section:generation-rule}, and that the auxiliary part of the tree coincides with the standard cut/all structure.
Since there are no transpositions or repetitions, these local statements determine the entire visited tree.
Therefore the search tree induced by Algorithm~\ref{tab:alg2} is exactly characterized by the node kinds $\{P,A',P',C,A\}$ and the generation rules in Section~\ref{section:generation-rule}.
\hfill$\square$

\subsection{Detailed Proof of Proposition~\ref{prop:tt-reuse}}
\label{app:proof-tt-reuse}
For this proof, interpret ``stored under obligation $\mathcal O_k$'' in the strong sense used in Section~\ref{sec:transposition-table}: the TT information currently available at the stored depth is sufficient to discharge the current-node requirements of $\mathcal O_k$ \emph{and} all recursively required child obligations (taking possible PV promotion into account).
Define a preorder $\succeq$ on obligations by
\[
\mathcal O_x \succeq \mathcal O_y
\quad\Longleftrightarrow\quad
\text{every TT package sufficient for }\mathcal O_x\text{ is also sufficient for }\mathcal O_y.
\]
Under this interpretation, the proposition supplies sufficient reuse tests based on producer obligations. A terminal node, a stronger certificate, or a combination of independently certified data may admit additional reuse.

We first recall the obligation content from Section~\ref{sec:obligations}.
A $P$-obligation requires an exact value, a canonical optimal move, no cutoff, the principal child under $P$, and each non-principal child under $A'$.
An $A'$-obligation requires an exact value, a canonical optimal move, the principal child under $P'$, and only alpha--beta refutations for the non-principal children.
A $P'$-obligation requires no cutoff and requires \emph{every} child to satisfy the $A'$-obligation.
Finally, $C$- and $A$-obligations require only the usual alpha--beta pruning information (valid lower/upper bounds, or an exact value) at sufficient depth.
By induction on the remaining play length, with the terminal case immediate, a $P$-certificate at a position is stronger than both an $A'$-certificate and a $P'$-certificate at the same position: it provides the current-node exact value and canonical move, forbids cutoff, and imposes recursively stronger child obligations than either $A'$ or $P'$ requires.

\emph{Sufficiency.}
We first show the positive inclusion statements.

\emph{(a) Query kind $P$.}
Trivially $\mathcal O_P\succeq \mathcal O_P$.
Therefore a previously stored $P$-entry is sufficient for a later $P$-query.

\emph{(b) Query kind $A'$.}
We have $\mathcal O_P\succeq \mathcal O_{A'}$ because a $P$-entry already provides an exact value and a canonical best move at the current node; moreover, its principal child is stored under $P$, which is stronger than the $P'$ requirement needed by $A'$, and its non-principal children are stored under $A'$, which is stronger than the mere refutations allowed at an $A'$-node.
Also $\mathcal O_{A'}\succeq \mathcal O_{A'}$ trivially.
Hence stored $P$- or $A'$-information is sufficient for an $A'$-query.

\emph{(c) Query kind $P'$.}
A $P'$-entry satisfies its own obligation. A $P$-entry covers every child: its principal $P$-child also satisfies $A'$, and its other children are $A'$-certified. Thus stored $P$- or $P'$-information suffices.

\emph{(d) Query kind $C$ or $A$.}
Only ordinary alpha--beta bound semantics matter. Exact values from any certified kind are usable, as are $C/A$ entries whose bound type (\textsc{Exact}, \textsc{Lower}, or \textsc{Upper}) and depth satisfy the current query. The producer kind is immaterial once these checks pass.

Weaker tags alone need not establish stronger obligations: $A'$ does not certify every child as $A'$, and $P'$ does not certify its principal child as $P'$. Additional stored evidence may justify further reuse. The sufficiency arguments establish Proposition~\ref{prop:tt-reuse}.

\hfill$\square$

\section{Game Rules}
\label{app:game-rules}
The following rules specify the two games used in Section~4. The terminal utilities remain in Sections~\ref{section:rule-othello} and~\ref{section:rule-connect4}.

\subsection{Othello}
\label{app:rules-othello}
\begin{enumerate}
\item The game is played on an \(N\times N\) board by two players, Black and White. Standard Othello uses \(N=8\); in our experiments we use \(N=6\).
\item The game starts with four discs in the center, arranged in a \(2\times 2\) square with opposite colors on the diagonals.
\item Players alternate turns, with Black moving first.
\item A move consists of placing a disc of the player's color on an empty square. Any opponent discs that are \emph{bracketed} between the newly placed disc and another disc of the same color, along any of the eight directions (horizontal, vertical, or diagonal), are flipped to the player's color.
\item A move is legal only if it flips at least one opponent disc.
\item If a player has no legal move, the player passes and the opponent moves. In the formal model of Section~\ref{sec:game-model}, such a forced pass is represented as the unique legal move.
\item The game ends when neither player has a legal move (equivalently, after two consecutive passes).
\item The winner is the player with more discs of their color on the board at the end of the game; if both players have the same number of discs, the game is a draw.
\end{enumerate}

\subsection{Connect Four}
\label{app:rules-connect4}
\begin{enumerate}
\item The game is played by two players on a \(7\times 6\) vertical grid (7 columns and 6 rows). We refer to the players as Red and Yellow.
\item Players alternate turns. On a turn, the player chooses a column that is not full and drops one disc into that column; the disc occupies the lowest empty cell in the chosen column.
\item A player wins immediately upon forming a line of four of their discs in a row, in any of the following directions: horizontal, vertical, or diagonal.
\item If the board becomes full (42 discs) without either player forming four in a row, the game is a draw.
\end{enumerate}

\section{Experimental Details}
\label{app:experimental-details}
This appendix gives the move-selection and deployment-policy details for the experiments in Section~4.

\subsection{Othello Move-Selection Searches and Ordering}
\label{app:othello-selectors}
The Othello counting convention distinguishes primary expansions from auxiliary move-selection expansions. The latter are excluded from the reported primary counts in Tables~\ref{tab:retro_result_66othello} and~\ref{tab:retro_result_66othello_full}.
\begin{itemize}
\item \textbf{Weak (alpha--beta).}
The reported counts measure positions expanded by the main alpha--beta/PVS procedure.
At cut nodes (fail-high nodes), we enforce that the first expanded move is guaranteed to fail high by running an auxiliary null-window probe search with window \([\beta-\epsilon,\beta]\) and selecting the move returned by that probe as the first move.
This auxiliary probe uses the same move-selection implementation and positional heuristic as described below, but the positions expanded by the probe are \emph{excluded} from Table~\ref{tab:retro_result_66othello}.

\item \textbf{Semi-strong (reopening alpha--beta).}
The reported counts measure the implementation's obligation-enforcing certification phase, including retained certified positions and auxiliary cut/all positions. This phase follows the node obligations described in Algorithm~\ref{tab:alg2-1}.
In the reported implementation, principal moves are selected before the corresponding certification expansion. At optimal-agent decision points this requires an optimal move; at auxiliary cut nodes it only requires a move that reaches the cutoff bound. The remaining moves, when required, follow the implementation's deterministic heuristic order.
To select an optimal move at an optimal-agent decision point, the implementation invokes an auxiliary alpha--beta/PVS search. Cut-node selectors instead use null-window fail-high probes. Positions expanded by these auxiliary selection searches are \emph{excluded} from Table~\ref{tab:retro_result_66othello}.
\end{itemize}

\paragraph{Move ordering heuristic (non-PV moves).}
For non-PV move ordering in the auxiliary searches and in the parts of the primary procedures where an explicit heuristic order is needed, we use a deterministic move-ordering heuristic combining:
(i) a square-dependent static weight (positional score),
(ii) the number of discs that become stable (cannot be flipped thereafter) immediately after making the move, and
(iii) mobility features, including the number of legal moves and the set of playable squares.
These features and their weights are engineering choices of this implementation. More generally, move ordering is central to alpha--beta efficiency~\cite{marsland1986pruning}.
The exact deterministic scoring function used for move ordering in the reported runs is given in the released implementation.
Its feature design and relative weighting were chosen manually for search engineering, informed by standard Othello-engine practice, and were not obtained by a systematic hyper-parameter optimization sweep.

\subsection{Stored Positions and a Fixed Deployment Policy}
\label{app:othello-policy}
Let \(S\) denote the retained position representatives carrying a certified kind \(P\), \(A'\), or \(P'\), under the counting conventions of Section~\ref{sec:othello-position-counts}. Within each search worker, a completed \(P\) or \(A'\) obligation is reused on subsequent optimal-agent visits, preserving its certified continuation. The postprocessing combines independently generated subproblems by merging their kind flags and bounds at each position.

A fixed deployment policy can be defined from the retained successor obligations: at a \(P\)-tagged position choose an optimal \(P\)-certified successor; otherwise, at an \(A'\)-tagged position choose an optimal \(P'\)- or \(P\)-certified successor. Resolve remaining choices by a fixed position-based order. Optimal terminal moves are eligible, and forced passes use the corresponding role and sign changes. Giving \(P\) priority handles combined kind flags, while \(P'\) obligations cover every free-agent continuation. By induction, the resulting policy's certified region is supported by \(S\), with pass and terminal handling implicit. The solution-artifact column reports \(|S|=97{,}252{,}321\) retained records after aggregation; it counts any additional certified records retained alongside the chosen policy's region.

\subsection{Connect Four Oracle Move Ordering}
\label{app:connect4-oracle}
Because we assume oracle access to exact WDL values, both the weak alpha--beta procedure and the semi-strong certification procedure order moves at every expanded position by decreasing value, and choose the first move from among the value-maximizing moves using the fixed center-first, then left-first tie-breaking rule of Section~4.2.1.
In particular, this holds at all node classes, including cut-node contexts: the first explored move is the move of \emph{maximum} value (ties broken deterministically), rather than merely some move that is guaranteed to fail high.
This oracle setting isolates structural coverage under value-optimal ordering and the stated tie-breaking rule. It is not a lower bound over all possible tie-breaking rules, compressed representations, or solution procedures.
Under oracle WDL access and the fixed center-first, then left-first tie-breaking rule, the procedures are fully deterministic; accordingly, each regime contributing to Table~\ref{tab:retro_result_connect4} was executed once, and the reported counts are those of that single deterministic run.

\clearpage
\section{Complete Position Counts}
\label{app:complete-counts}
Tables~\ref{tab:retro_result_66othello_full} and~\ref{tab:retro_result_connect4_full} give all disc-count rows underlying the shortened main-text Tables~\ref{tab:retro_result_66othello} and~\ref{tab:retro_result_connect4}. The counting conventions are those of Sections~\ref{sec:othello-position-counts} and~\ref{sec:connect4-position-counts}. In every table, the total includes all disc counts, not just the selected rows shown in the main text.

\begingroup\small
\begin{longtable}{r|r|r|r|r|r}
\caption{Complete distinct-position counts for $6\times6$ Othello, using the conventions of Section~\ref{sec:othello-position-counts}. Auxiliary move-selection searches are excluded from primary counts; forced-pass-only positions are excluded throughout. The solution artifact is the retained certified set $S$; the certificate column contains auxiliary cut/all positions. Daggers mark unfinished BFS enumeration.}
\label{tab:retro_result_66othello_full}\\\hline
category & weak & \multicolumn{3}{c|}{semi-strong} & strong\\\hline
regime & alpha--beta & \multicolumn{3}{c|}{reopening alpha--beta} & exhaustive BFS\\\hline
component & (total) & solution artifact & proof certificate & total & (total)\\\hline\hline
\multicolumn{1}{r|}{discs} & \multicolumn{5}{c}{number of positions} \\\hline
\endfirsthead

\hline
category & weak & \multicolumn{3}{c|}{semi-strong} & strong\\\hline
regime & alpha--beta & \multicolumn{3}{c|}{reopening alpha--beta} & exhaustive BFS\\\hline
component & (total) & solution artifact & proof certificate & total & (total)\\\hline\hline
\multicolumn{1}{r|}{discs} & \multicolumn{5}{c}{number of positions} \\\hline
\endhead

\hline
\endfoot
\endlastfoot

4 & 1 & 1 & 0 & 1 & 1 \\
5 & 1 & 1 & 0 & 1 & 1 \\
6 & 3 & 3 & 0 & 3 & 3 \\
7 & 7 & 7 & 7 & 14 & 14 \\
8 & 12 & 12 & 20 & 32 & 60 \\
9 & 32 & 32 & 86 & 118 & 314 \\
10 & 59 & 63 & 233 & 296 & 1,632 \\
11 & 151 & 163 & 771 & 934 & 9,069 \\
12 & 287 & 343 & 2,121 & 2,464 & 51,964 \\
13 & 731 & 869 & 6,223 & 7,092 & 292,946 \\
14 & 1,382 & 1,842 & 17,689 & 19,531 & 1,706,168 \\
15 & 3,549 & 4,506 & 47,270 & 51,776 & 9,289,258 \\
16 & 6,864 & 10,015 & 132,731 & 142,746 & 51,072,917 \\
17 & 17,812 & 22,856 & 340,377 & 363,233 & 251,070,145 \\
18 & 35,006 & 50,594 & 923,911 & 974,505 & 1,208,692,475 \\
19 & 90,240 & 106,165 & 2,295,318 & 2,401,483 & 5,014,312,131 \\
20 & 177,787 & 225,426 & 6,045,608 & 6,271,034 & 19,791,417,568 \\
21 & 447,687 & 427,842 & 14,395,120 & 14,822,962 & 65,844,387,711 \\
22 & 858,184 & 839,607 & 36,014,638 & 36,854,245 & 203,504,012,437 \\
23 & 2,080,872 & 1,437,738 & 80,463,765 & 81,901,503 & 525,923,099,578 \\
24 & 3,815,862 & 2,531,925 & 188,487,830 & 191,019,755 & 1,234,638,103,732 \\
25 & 8,819,433 & 3,896,228 & 390,330,164 & 394,226,392 & 2,417,685,025,700 \\
26 & 15,263,415 & 6,035,411 & 832,468,650 & 838,504,061 & $\dagger$ \\
27 & 32,755,025 & 8,219,315 & 1,545,648,954 & 1,553,868,269 & $\dagger$ \\
28 & 52,251,880 & 10,922,915 & 2,846,423,962 & 2,857,346,877 & $\dagger$ \\
29 & 99,391,021 & 12,757,435 & 4,463,985,509 & 4,476,742,944 & $\dagger$ \\
30 & 139,488,250 & 13,919,043 & 6,520,602,455 & 6,534,521,498 & $\dagger$ \\
31 & 215,616,210 & 13,143,268 & 7,811,810,149 & 7,824,953,417 & $\dagger$ \\
32 & 240,828,705 & 10,736,694 & 7,894,710,441 & 7,905,447,135 & $\dagger$ \\
33 & 255,621,773 & 7,156,567 & 6,022,021,827 & 6,029,178,394 & $\dagger$ \\
34 & 180,738,857 & 3,550,378 & 3,229,064,205 & 3,232,614,583 & $\dagger$ \\
35 & 90,900,623 & 1,255,057 & 1,003,741,866 & 1,004,996,923 & $\dagger$ \\\hline
total & 1,339,211,721 & 97,252,321 & 42,889,981,900 & 42,987,234,221 & $\gg$ 4,473,922,545,824 \\\hline
\end{longtable}
\endgroup

\clearpage

\begingroup\small
\begin{longtable}{r|r|r|r|r|r}
\caption{Complete distinct-position counts for Connect Four, using the conventions of Section~\ref{sec:connect4-position-counts}. Weak and semi-strong counts use oracle WDL ordering and fixed tie-breaking. Terminal positions are included, and reflected positions are counted separately. The strong column gives published rule-reachable counts~\cite{boeck2025connect4strong}.}
\label{tab:retro_result_connect4_full}\\\hline
category & weak & \multicolumn{3}{c|}{semi-strong} & strong\\\hline
regime & alpha--beta & \multicolumn{3}{c|}{reopening alpha--beta} & rule-reachable\\\hline
component & (total) & solution artifact & proof certificate & total & (total)\\\hline\hline
\multicolumn{1}{r|}{discs} & \multicolumn{5}{c}{number of positions} \\\hline
\endfirsthead

\hline
category & weak & \multicolumn{3}{c|}{semi-strong} & strong\\\hline
regime & alpha--beta & \multicolumn{3}{c|}{reopening alpha--beta} & rule-reachable\\\hline
component & (total) & solution artifact & proof certificate & total & (total)\\\hline\hline
\multicolumn{1}{r|}{discs} & \multicolumn{5}{c}{number of positions} \\\hline
\endhead

\endfoot
\endlastfoot

0 & 1 & 1 & 0 & 1 & 1 \\
1 & 1 & 7 & 0 & 7 & 7 \\
2 & 7 & 13 & 12 & 25 & 49 \\
3 & 7 & 49 & 12 & 61 & 238 \\
4 & 47 & 87 & 95 & 182 & 1,120 \\
5 & 47 & 274 & 90 & 364 & 4,263 \\
6 & 260 & 477 & 534 & 1,011 & 16,422 \\
7 & 257 & 1,350 & 520 & 1,870 & 54,859 \\
8 & 1,082 & 2,117 & 2,629 & 4,746 & 184,275 \\
9 & 1,056 & 5,349 & 2,524 & 7,873 & 558,186 \\
10 & 3,747 & 7,616 & 10,050 & 17,666 & 1,662,623 \\
11 & 3,506 & 18,928 & 9,245 & 28,173 & 4,568,683 \\
12 & 12,442 & 25,741 & 34,093 & 59,834 & 12,236,101 \\
13 & 10,924 & 61,889 & 30,369 & 92,258 & 30,929,111 \\
14 & 39,826 & 82,460 & 105,072 & 187,532 & 75,437,595 \\
15 & 34,201 & 191,105 & 90,730 & 281,835 & 176,541,259 \\
16 & 118,507 & 247,668 & 293,905 & 541,573 & 394,591,391 \\
17 & 100,607 & 535,663 & 248,031 & 783,694 & 858,218,743 \\
18 & 313,016 & 667,698 & 721,836 & 1,389,534 & 1,763,883,894 \\
19 & 264,026 & 1,345,153 & 601,283 & 1,946,436 & 3,568,259,802 \\
20 & 736,986 & 1,618,273 & 1,573,007 & 3,191,280 & 6,746,155,945 \\
21 & 621,892 & 3,037,766 & 1,310,651 & 4,348,417 & 12,673,345,045 \\
22 & 1,547,226 & 3,528,785 & 3,051,976 & 6,580,761 & 22,010,823,988 \\
23 & 1,309,926 & 6,100,604 & 2,557,002 & 8,657,606 & 38,263,228,189 \\
24 & 2,857,888 & 6,816,505 & 5,246,286 & 12,062,791 & 60,830,813,459 \\
25 & 2,440,699 & 10,766,849 & 4,444,052 & 15,210,901 & 97,266,114,959 \\
26 & 4,598,788 & 11,557,178 & 7,927,823 & 19,485,001 & 140,728,569,039 \\
27 & 3,969,020 & 16,729,937 & 6,799,724 & 23,529,661 & 205,289,508,055 \\
28 & 6,473,206 & 17,250,960 & 10,560,070 & 27,811,030 & 268,057,611,944 \\
29 & 5,651,727 & 22,868,076 & 9,172,217 & 32,040,293 & 352,626,845,666 \\
30 & 7,927,180 & 22,627,524 & 12,442,361 & 35,069,885 & 410,378,505,447 \\
31 & 7,034,569 & 27,633,942 & 10,974,975 & 38,608,917 & 479,206,477,733 \\
32 & 8,489,314 & 26,253,679 & 13,040,374 & 39,294,053 & 488,906,447,183 \\
33 & 7,646,337 & 29,211,163 & 11,659,117 & 40,870,280 & 496,636,890,702 \\
34 & 7,706,523 & 26,379,767 & 11,842,333 & 38,222,100 & 433,471,730,336 \\
35 & 7,000,444 & 26,442,273 & 10,688,538 & 37,130,811 & 370,947,887,723 \\
36 & 5,843,979 & 22,724,558 & 9,178,727 & 31,903,285 & 266,313,901,222 \\
37 & 5,397,649 & 19,532,791 & 8,429,219 & 27,962,010 & 183,615,682,381 \\
38 & 3,259,327 & 15,488,457 & 5,287,756 & 20,776,213 & 104,004,465,349 \\
39 & 3,080,195 & 10,458,832 & 4,964,166 & 15,422,998 & 55,156,010,773 \\
40 & 983,599 & 7,332,236 & 1,818,241 & 9,150,477 & 22,695,896,495 \\
41 & 970,631 & 2,897,044 & 1,748,096 & 4,645,140 & 7,811,825,938 \\
42 & 0 & 1,683,726 & 441,663 & 2,125,389 & 1,459,332,899 \\\hline
total & 96,450,672 & 342,134,570 & 157,309,404 & 499,443,974 & 4,531,985,219,092 \\\hline
\end{longtable}
\endgroup

\section{Online Resources}
\label{sec:online-resources}
\begin{itemize}
\item \url{https://zenodo.org/records/18843225}
\item \url{https://github.com/eukaryo/reopening-alphabeta-experiment/}
\item \url{https://github.com/eukaryo/connect4-semi-strong-experiment/}
\item Connect Four upstream strong-solution dataset used by the public reproduction code: \url{https://doi.org/10.5281/zenodo.14582823}
\end{itemize}

\end{document}

%% file: publication_metadata.tex
\acmDOI{10.1613/jair.1.22432}
\acmVolume{87}
\acmArticle{29}
\acmMonth{11}
\acmYear{2026}
\received{03 April 2026}
\received[accepted]{05 November 2026}